\documentclass{article}

\usepackage{microtype}
\usepackage{graphicx}
\usepackage{subfigure}
\usepackage{multirow}
\usepackage{booktabs} %
\usepackage{float}

\usepackage{hyperref}

\usepackage[accepted]{icml2024}

\usepackage{amsmath}
\usepackage{amssymb}
\usepackage{mathtools}
\usepackage{amsthm}

\usepackage[capitalize,noabbrev]{cleveref}

\theoremstyle{plain}

\theoremstyle{definition}

\theoremstyle{remark}

\usepackage[textsize=tiny]{todonotes}

\usepackage{hyperref}
\usepackage{url}

\usepackage{microtype}
\usepackage{graphicx}
\usepackage{subfigure}
\usepackage{booktabs} %
\usepackage{subfigure}
\usepackage{amsmath}
\usepackage{amssymb}
\usepackage{mathtools}
\usepackage{amsthm}
\usepackage{wrapfig}
\usepackage{booktabs}
\usepackage{bm}
\usepackage{color}
\usepackage{marvosym}

\def\1{\bm{1}}
\newcommand{\err}[1]{\scriptsize $\pm$ #1}

\icmltitlerunning{Efficient Adaptation in Mixed-Motive Environments via Hierarchical Opponent Modeling and Planning}

\begin{document}

\twocolumn[
\icmltitle{Efficient Adaptation in Mixed-Motive Environments \\
via Hierarchical Opponent Modeling and Planning}

\icmlsetsymbol{equal}{*}

\begin{icmlauthorlist}
\icmlauthor{Yizhe Huang}{pku,bigai}
\icmlauthor{Anji Liu}{ucla}
\icmlauthor{Fanqi Kong}{bigai,thu}
\icmlauthor{Yaodong Yang}{pku,bigai}
\icmlauthor{Song-Chun Zhu}{pku,bigai}
\icmlauthor{Xue Feng\textsuperscript{\Letter}}{bigai}
\end{icmlauthorlist}

\icmlaffiliation{pku}{Institute for Artificial Intelligence, Peking University}
\icmlaffiliation{ucla}{University of California, Los Angeles}
\icmlaffiliation{thu}{Tsinghua University}
\icmlaffiliation{bigai}{State Key Laboratory of General Artificial Intelligence, BIGAI}

\icmlcorrespondingauthor{Xue Feng}{fengxue@bigai.ai}

\icmlkeywords{Machine Learning, ICML}

\vskip 0.3in
]

\printAffiliationsAndNotice{}  %

\begin{abstract}
Despite the recent successes of multi-agent reinforcement learning (MARL) algorithms, efficiently adapting to co-players in mixed-motive environments remains a significant challenge.
One feasible approach is to hierarchically model co-players' behavior based on inferring their characteristics. However, these methods often encounter difficulties in efficient reasoning and utilization of inferred information.
To address these issues, we propose Hierarchical Opponent modeling and Planning (HOP), a novel multi-agent decision-making algorithm that enables few-shot adaptation to unseen policies in mixed-motive environments. HOP is hierarchically composed of two modules: an opponent modeling module that infers others' goals and learns corresponding goal-conditioned policies, and a planning module that employs Monte Carlo Tree Search (MCTS) to identify the best response.
Our approach improves efficiency by updating beliefs about others' goals both across and within episodes and by using information from the opponent modeling module to guide planning.
Experimental results demonstrate that in mixed-motive environments, HOP exhibits superior few-shot adaptation capabilities when interacting with various unseen agents, and excels in self-play scenarios. Furthermore, the emergence of social intelligence during our experiments underscores the potential of our approach in complex multi-agent environments.
\end{abstract}

\section{Introduction}
\label{Introduction}

Constructing agents being able to rapidly adapt to previously unseen agents is a longstanding challenge for Artificial Intelligence. We refer to this ability as few-shot adaptation. Previous work has proposed well-performed MARL algorithms to study few-shot adaptation in zero-sum games~\cite{vinyals2019grandmaster, vezhnevets2020options} and common-interest environments~\cite{barrett2011empirical, hu2020other, mahajan2022generalization, mirsky2022survey, bauer2023human}. These environments involve a predefined competitive or cooperative relationship between agents. However, the majority of realistic multi-agent decision-making scenarios are not confined to these situations and should be abstracted as mixed-motive environments~\cite{komorita1995interpersonal, dafoe2020open}, where the relationships between agents are non-deterministic, and the best responses of an agent may change with others' behavior.
A policy, that is unable to quickly adapt to co-players, may harm not only the focal agent's interest but also the entire group's benefit.
Therefore, fast adapting to new co-players in mixed-motive environments warrants significant attention, but there has been little focus on this aspect. 

In this paper, we focus on the few-shot adaptation to unseen agents in mixed-motive environments.
Many algorithms struggle to perform well in mixed-motive environments despite success in zero-sum and pure-cooperative environments, because they use efficient techniques specific to reward structures, such as minimax~\cite{littman1994markov, li2019robust}, Double Oracle~\cite{mcmahan2003planning, balduzzi2019open} or IGM condition~\cite{sunehag2017value, son2019qtran, rashid2020monotonic}, which are not applicable in mixed-motive environments. 
The non-deterministic relationships between agents and the general-sum reward structure make decision-making and few-shot adaptation more challenging in mixed-motive environments compared with zero-sum and pure-cooperative environments.

According to cognitive psychology and related disciplines, humans' ability to rapidly solve previously unseen problems depends on hierarchical cognitive mechanisms~\cite{butz2016mind, kleiman2016coordinate, eppe2022intelligent}. 
This hierarchical structure unifies high-level goal reasoning with low-level action planning. 
Meanwhile, research on machine learning also emphasizes the importance and effectiveness of hierarchical goal-directed planning for few-shot problem-solving~\cite{eppe2022intelligent}. 
Inspired by the hierarchical structure, we propose an algorithm, named Hierarchical Opponent modeling and Planning (HOP), for tackling few-shot adaptation in mixed-motive environments.
HOP hierarchically consists of two modules: an opponent modeling module and a planning module.
The opponent modeling module infers co-players' goals and learns their goal-conditioned policies, based on Theory of Mind (ToM) - the ability to understand others' mental states (like goals and beliefs) from their actions~\cite{baker2017rational}.
More specifically, to improve inference efficiency, beliefs about others' goals are updated both between and within episodes.
Then, the information from the opponent modeling module is sent to the planning module, which is based on Monte Carlo Tree Search (MCTS), to compute the next action.

To assess the few-shot adaptation ability of HOP, we conduct experiments in Markov Stag-Hunt (MSH) and Markov Snowdrift Game (MSG), which spatially and temporally extend two classic paradigms in game theory: the Stag-Hunt game~\cite{rousseau1999discourse} and the Snowdrift game(also known as the game of chicken or hawk-dove game)~\cite{rapoport1966game}.
Both of the two games illustrate how the best response in a mixed-motive environment is influenced by the strategy of co-players. 
Experimental results illustrate that in these environments, HOP exhibits superior few-shot adaptation ability compared with baselines, including the well-established MARL algorithms LOLA, social influence, A3C, prosocial-A3C, PR2, and a model-based algorithm direct-OM. Meanwhile, HOP achieves high rewards in self-play, showing its exceptional decision-making ability in mixed-motive games. 
In addition, we observe the emergence of social intelligence from the interaction between multiple HOP agents, such as self-organized cooperation and alliance of the disadvantaged.

\section{Related Work}

MARL has explored multi-agent decision-making in mixed-motive games. One approach is to add intrinsic rewards to incentivize collaboration and consideration of the impact on others, alongside maximizing extrinsic rewards. Notable examples include ToMAGA~\cite{nguyen2020theory}, MARL with inequity aversion~\cite{hughes2018inequity}, and prosocial MARL~\cite{peysakhovich2018prosocial}. However, many of these algorithms rely on hand-crafted intrinsic rewards and assume access to rewards of co-players, which can make them exploitable by self-interested algorithms and less effective in realistic scenarios where others' rewards are not visible~\cite{komorita1995interpersonal}. To address these issues, \citet{jaques2019social} have included intrinsic social influence reward that use counterfactual reasoning to assess the effect of an agent's actions on its co-players' behavior.

LOLA~\cite{foerster2018learning} and its extension (such as POLA~\cite{zhao2022proximal}, M-FOS~\cite{lu2022model}) consider the impact of one agent's learning process, rather than treating them as a static part of the environment. However, LOLA requires knowledge of co-players' network parameters, which may not be feasible in many scenarios. LOLA with opponent modeling relaxes this requirement, but scaling problems may arise in complex sequential environments that require long action sequences for rewards.

Our work relates to opponent modeling (see~\cite{albrecht2018autonomous} for a comprehensive review).
I-POMDP~\cite{gmytrasiewicz2005framework} is a typical opponent modeling and planning framework, which maintains dynamic beliefs over the physical environment and beliefs over co-players' beliefs.
It maximizes a value function of the beliefs to determine the next action. However, the nested belief inference suffers from serious computational complexity problems, which makes it impractical in complex environments. 
Unlike I-POMDP and its approximation methods~\cite{doshi2008generalized, doshi2009monte, hoang2013interactive,  han2018learning, han2019ipomdp, zhang2022sipomdplite}, HOP explicitly uses beliefs over co-players' goals and policies to learn a neural network model of co-players, which guides an MCTS planner to compute next actions.
HOP avoids nested belief inference and performs sequential decision-making more efficiently.

Theory of mind (ToM), originally a concept of cognitive science and psychology~\cite{baron1985does}, has been transformed into computational models over the past decade and used to infer agents' mental states such as goals and desires. Bayesian inference has been a popular technique used to make ToM computational~\cite{baker2011bayesian, poppel2018satisficing, wu2021too, zhi2022solving}. With the rapid development of the neural network, some recent work has attempted to achieve ToM using neural networks~\cite{rabinowitz2018machine, shu2018m, wen2019probabilistic,moreno2021neural}. 
HOP gives a practical and effective framework to utilize ToM, and extend its application scenarios to mixed-motive environments, where both competition and cooperation are involved and agents' goals are private and volatile.

Monte Carlo Tree Search (MCTS) is a widely adopted planning method for optimal decision-making. Recent work, such as AlphaZero~\cite{silver2018general} and MuZero~\cite{schrittwieser2020mastering} have used MCTS as a general policy improvement operator over the base policy learned by neural networks. However, MCTS is limited in multi-agent environments, where the joint action space grows rapidly with the number of agents~\cite{choudhury2022scalable}. We avoid this problem by estimating the policies of co-players and planning only for the focal agent's actions.

BAMDP~\cite{duff2002optimal} is a principled framework for handling uncertainty in dynamic environments. It maintains a posterior distribution over the transition probabilities, which is updated using Bayes' rule as new data becomes available. Several algorithms~\cite{guez2012efficient, zintgraf2019varibad, rigter2021risk} have been developed based on BAMDP, but they are designed for single-agent environments. BA-MCP~\cite{guez2012efficient} employs the Monte Carlo Tree Search (MCTS) method to provide a sample-based approach grounded in BAMDP. However, it assumes a fixed transition function distribution to be learned interactively, posing challenges in multi-agent scenarios due to the co-player's strategy under an unknown distribution. \cite{ng2012bayes} combines BAMDP with I-POMDP in an attempt to address multi-agent problems. However, this integration introduces computational complexity issues similar to those of I-POMDP, as previously discussed. In contrast, HOP efficiently handles both reward and transition uncertainties, and extends MCTS to multi-agent scenarios, offering a scalable solution for multi-agent environments.

Numerous real-world scenarios, including autonomous driving, human-machine interaction and multi-player sports, can be effectively modeled as mixed-motive games. Existing research~\cite{fisac2018probabilistically, nakamura2023online, pmlr-v229-hu23b} has explored planning and controlling robots in these real multi-agent environments, relying on predictions of other agents' behavior within the scene. These studies primarily concentrate on robot control within specific scenarios. In contrast, our environment abstracts the mixed motivation factors inherent in these scenarios, enabling representation of a broader range of scenarios and facilitating the development of more general algorithms. We believe HOP holds significant potential for application in various real-life scenarios.

\section{Problem Formulation}
We consider multi-agent hierarchical decision-making in mixed-motive environments, which can be described as a Markov game~\cite{littman1994markov} with goals, specified by a tuple $<N, S, \mathbf{A}, T, \mathbf{R}, \gamma, T_{max}, \mathbf{G}>$.

Here, agent $i \in N = \{1, 2, \cdots, n\}$ chooses action from action space $A_i =\{a_i\}$. $\mathbf{A} = A_1 \times A_2 \times \cdots \times A_n$  is the joint action space. 
The joint action $\bm{a}_{1:n} \in \mathbf{A}$ will lead to a state transition based on the transition function $T: S\times \mathbf{A} \times S \rightarrow [0, 1]$. 
Specifically, after agents take the joint action $\bm{a}_{1:n}$ the state of the environment will transit from $s$ to $s'$ with probability $T(s' | s, \bm{a}_{1:n})$.
The reward function $R_i: S \times \mathbf{A} \rightarrow \mathbb{R}$ denotes the immediate reward received by agent $i$ after joint action $\bm{a}_{1:n}$ is taken on state $s \in S$.
The discount factor for future rewards is denoted as $\gamma$.
$T_{max}$ is the maximum length of an episode. 
$\pi_i: S \times A_i \rightarrow [0, 1]$ denotes agent $i$'s policy, specifying the probability $\pi_i(a_i | s)$ that agent $i$ chooses action $a_i$ at state $s$.

The environments we study have a set of goals, denoted by $\mathbf{G} = G_1 \times G_2 \times \cdots \times G_n$, where $G_i = \{g_{i,1}, \cdots, g_{i,|G_i|}\}$ represents the set of goals for agent $i$. 
$g_{i, k}$ is a set of states, where $g_{i, k} \cap g_{i, k'} = \emptyset, \forall\ k \neq k'$. We would say agent $i$'s goal is $g_{i, k_0}$ at time $t$, if $\exists t' \geq 0, s^{t+t'} \in g_{i, k_0}$ and $\forall\ 0 \leq t'' < t', 0 \leq k \leq |G_i|, s^{t+t''} \notin g_{i, k}$.
For any two agents $i$ and $j$, $i$ can infer $j$'s goal based on its trajectory. 
Specifically, $i$ maintains a belief over $j$'s goals, $b_{ij}: G_j \rightarrow [0, 1]$, which is a probability distribution over $G_j$.

Here, algorithms are evaluated in terms of self-play and few-shot adaptation to unseen policies in mixed-motive environments. Self-play involves multiple agents using the same algorithm to undergo training from scratch. The performance of algorithms in self-play is evaluated by their expected reward after convergence. Self-play performance demonstrates the algorithm's ability to make autonomous decisions in mixed-motive environments. 
Few-shot adaptation refers to the capability to recognize and respond appropriately to unknown policies within a limited number of episodes. The performance of algorithms in few-shot adaptation is measured by the rewards they achieve after engaging in these brief interactions.

\section{Methodology}
In this section, we propose \textbf{H}ierarchical \textbf{O}pponent modeling and \textbf{P}lanning (HOP), a novel algorithm for multi-agent decision-making in mixed-motive environments. HOP consists of two main modules: an opponent modeling module to infer co-players' goals and predict their behavior and a planning module to plan the focal agent's best response guided by the inferred information from the opponent modeling module.

\begin{figure*}[t]
  \centering
  \includegraphics[width=0.9\linewidth]{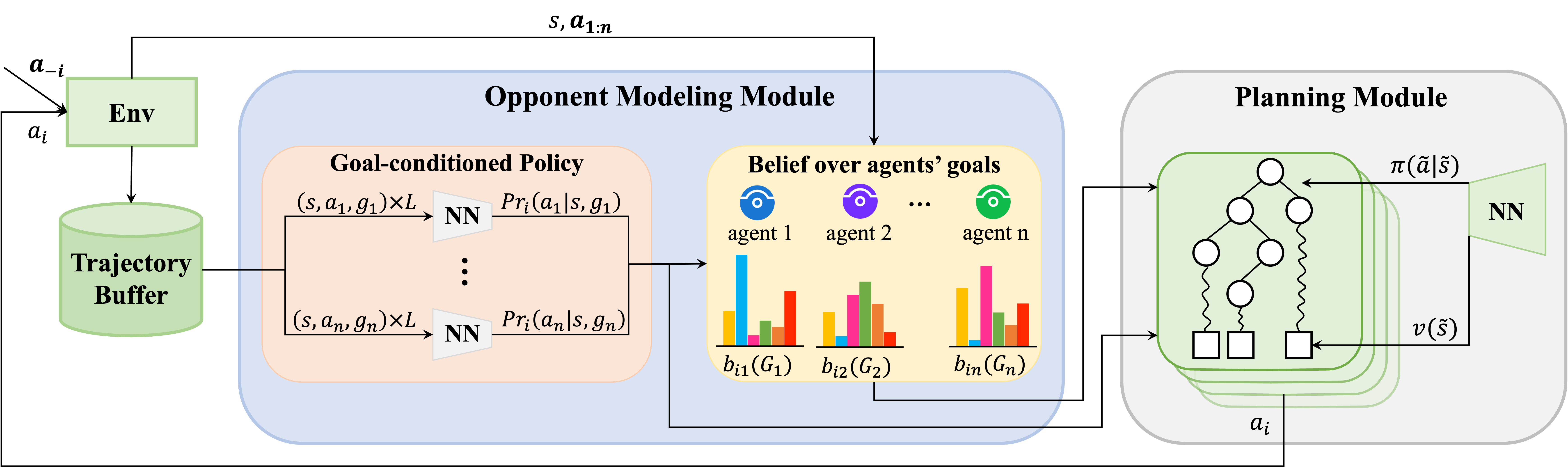}
  \caption{Overview of HOP. HOP consists of an opponent modeling module and a planning module. The opponent modeling module models the behavior of co-players by inferring co-players' goals and learning their goal-conditioned policies. Estimated behavior is then fed to the planning module to select a rewarding action for the focal agent.}
  \label{fig:method}
\end{figure*}

Based on the hypothesis in cognitive psychology that agents' behavior is goal-directed~\cite{gergely1995taking, buresh2007infants}, and that agents behave stably for a specific goal~\cite{warren2006dynamics}, the opponent modeling module models behavior of co-players with two levels of hierarchy. At the high-level, the module infers co-players' internal goals by analyzing their action sequences. Based on the inferred goals and the current state of the environment, the low-level component learns goal-conditioned policies to model the atomic actions of co-players.

In the planning module, MCTS is used to plan for the best response of the focal agent based on the inferred co-players' policies. To handle the uncertainty over co-players' goals, we sample multiple goal combinations of all co-players from the current belief and return the action that maximizes the average return over the sampled configurations. Following AlphaZero~\cite{silver2018general} and MuZero~\cite{schrittwieser2020mastering}, we maintain a policy and a value network to boost MCTS planning and in turn use the planned action and its value to update the neural network.

\cref{fig:method} gives an overview of HOP, and the pseudo-code of HOP is provided in \cref{sec: pseudo code}.

\subsection{Opponent Modeling with Efficient Adaptation}
\label{sec:opponent-modeling}

In goal-inference (as the light yellow component shown in Figure 1), HOP summarizes the co-players' objectives based on the interaction history. However, it faces the challenge of the co-player's goals potentially changing within episodes. To solve these issues, we propose two update procedures based on ToM: intra opponent modeling (intra-OM), which infers the co-player's immediate goals within a single episode, and inter opponent modeling (inter-OM), which summarizes the co-player's goals based on their historical episodes.
Intra-OM reasons about the goal of co-player $j$ in the current episode $K$ according to $j$'s past trajectory in episode $K$. It ensures that HOP is able to quickly respond to in-episode behavior changes of co-players. Specifically, in episode $K$, agent $i$'s belief about agent $j$'s goals at time $t$, $b_{ij}^{K,t}(g_j)$, is updated according to: 
\begin{align}
\begin{split}
    b_{ij}^{K, t+1}(g_j) &= Pr(g_j\ | s^{K,0:t+1}, a_j^{K,0:t}) \\
    &= Pr(g_j | s^{K, 0:t}, a_j^{K, 0:t-1}) \\
    & \ \ \ \ \  \cdot Pr_i(a_j^{K, t} | s^{K, 0:t}, a_j^{K, 0:t-1}, g_j)  \\
    & \ \ \ \ \  \cdot \frac{Pr(s^{K, t+1} | s^{K, 0:t}, a_j^{K, 0:t}, g_j)}{Pr_i(s^{K, t+1}, a_j^{K, t} | s^{K, 0:t}, a_j^{K, 0:t-1})}  \\
    &= \frac{1}{Z_1} b_{ij}^{K, t}(g_j) Pr_i(a_j^{K, t}|s^{K, t}, g_j),
\end{split}
    \label{eq: intra}
\end{align}
where we follow the Markov assumption $Pr(s^{K, t+1} | s^{K, 0:t}, a_j^{K, 0:t}, g_j) = Pr(s^{K, t+1} | s^{K,t}, a_j^{K, t})$ and model the co-player $j$ to maintain a Markov policy $Pr_i(a_j^{K, t}|s^{K, t}, g_j) = Pr_i(a_j^{K, t}|s^{K, 0:t}, g_j)$, and $Z_1 = \frac{Pr_i(s^{K, t+1}, a_j^{K, t} |s^{K, t})}{Pr(s^{K, t+1} | s^{K,t}, a^{K, t})}$ is the normalization factor that makes $\sum_{g_j \in G_j} {b_{ij}^{K,t+1}(g_j)}=1$. The likelihood term $Pr_i(a_j^{K, t}|s^{K, t}, g_j)$ is provided by the estimated goal-conditioned policies of co-players, which are described in the following.

However, intra-OM may suffer from inaccuracy of the prior (i.e., $b_{ij}^{K,0}(g_j)$) when past trajectories are not long enough for updates. Inter-OM makes up for this by calculating a precise prior based on past episodes. Belief update between two adjacent episodes is defined as:
\begin{equation}
    b_{ij}^{K,0}(g_j) = \frac{1}{Z_2} [\alpha b_{ij}^{K-1,0}(g_j) + (1 - \alpha) \1(g_j^{K-1} = g_j)],
    \label{eq: inter}
\end{equation}
where $\alpha \in [0, 1]$ is the horizon weight, which controls the importance of the history. As $\alpha$ decreases, agents attach greater importance to recent episodes. $\1(\cdot)$ is the indicator function. $Z_2$ is the normalization factor. The equation is equivalent to a time-discounted modification of the Monte Carlo estimate. Inter-OM summarizes co-players' goals according to all the previous episodes, which is of great help when playing with the same agents in a series of episodes.

The goal-conditioned policy (as the light orange component shown in \cref{fig:method}) $\pi_{\boldsymbol{\omega}}(a_j^{K, t}|s^{K, t}, g_j)$ is obtained through a neural network $\boldsymbol{\omega}$.
To train the network, a set of $(s^{K,t}, a_j^{K,t}, g_j^{K,t})$ is collected from episodes and sent to the replay buffer. $\boldsymbol{\omega}$ is updated at intervals to minimize the negative log-likelihood:
\begin{equation}
    L(\boldsymbol{\omega}) \!=\! \mathbb{E}[-\log(\pi_{\boldsymbol{\omega}}(a_j^{K, t}|s^{K, t}, g_j^{K, t}))].
    \label{eq: modeling loss}
\end{equation}

\subsection{Planning under Uncertain Co-player Models}
\label{sec:planning}
Given the policies of co-players estimated by the opponent modeling module, we can leverage planning algorithms such as MCTS to compute an advantageous action. However, a key obstacle to applying MCTS is that co-player policies estimated by the opponent modeling module contain uncertainty over co-players' goals. Naively adding such uncertainty as part of the environment would add a large bias to the simulation and degrade planning performance. To overcome this problem, we propose to sample co-players’ goal combinations according to the belief maintained by the opponent modeling module, and then estimate action value by MCTS based on the samples. To balance the trade-off between computational complexity and planning performance, we repeat the process multiple times and choose actions according to the average action value. In the following, we first introduce the necessary background of MCTS. We then proceed to introduce how we plan for a rewarding action under the uncertainty over co-player policies. 

\paragraph{MCTS} Monte Carlo Tree Search (MCTS) is a type of tree search that plans for the best action at each time step~\cite{silver2010monte, liu2018watch}. MCTS uses the environment to construct a search tree (right side of \cref{fig:method}) where nodes correspond to states and edges refer to actions. Specifically, each edge transfers the environment from its parent state to its child state. MCTS expands the search tree in ways (such as pUCT) that properly balance exploration and exploitation. Value and visit of every state-action (node-edge) pair are recorded during expansion~\cite{silver2016mastering}. Finally, the action with the highest value (or highest visit) of the root state (node) is returned and executed in the environment.

\paragraph{Planning under uncertain co-player policies} Based on beliefs over co-players' goals and their goal-conditioned policies from the opponent modeling module, we run MCTS for $N_s$ rounds. In each round, co-players' goals are sampled according to the focal agent's belief over co-players' goals $b_{ij}(g_j)$. Specifically, at time $t$ in episode $K$, we sample the goal combination $\mathbf{g}_{-i}=\{ g_j \sim b_{ij}^{K,t}(\cdot), j \neq i\}$. Then at every state $\tilde{s}^k$ in the MCTS tree of this round, co-players' actions $\mathbf{\tilde{a}}_{-i}$ are determined by $\mathbf{\tilde{a}}_{-i} \sim \pi_{\boldsymbol{\omega}}(\cdot | \tilde{s}^k, \mathbf{g}_{-i})$ from the goal-conditioned policy.

In each round, MCTS gives the estimated action value of the current state $Q(s^{K,t}, a, \mathbf{g}_{-i}) = V(\tilde{s'}(a)) $ $(a \in A_i)$, where $\tilde{s'}(a)$ is the next state after taking $\tilde{\mathbf{a}}_{-i}^0 \cup a$ from $\tilde{s}^0 = s^{K,t}$.

We average the estimated action value from MCTS in all $N_s$ rounds:
\begin{equation}
    Q_{avg}(s^{K,t}, a)=\sum\nolimits_{l=1}^{N_s} Q_l(s^{K,t}, a, \mathbf{g}_{-i}^l).
    \label{eq: Qavg}
\end{equation}
Agent $i$’s policy follows Boltzmann rationality model~\cite{baker2017rational}:
\begin{equation}
    \pi_{MCTS}(a|s^{K,t})=\frac{\exp ( \beta Q_{avg}(s^{K,t}, a))}{\sum_{a' \in A_i}\exp (\beta Q_{avg}(s^{K,t}, a'))},
    \label{eq: pi_MCTS}
\end{equation}
where $\beta \in [0, \infty)$ is rationality coefficient. As $\beta$ increases, the policy gets more rational. We choose our action at time $t$ of the episode $K$ based on $\pi_{MCTS}(a|s^{K,t})$.

Note that the effectiveness of MCTS is highly associated with the default policies and values provided to MCTS. 
When they are close to the optimal ones, they can offer an accurate estimate of state value, guiding MCTS search in the right direction. 
Therefore, following \citet{silver2018general}, we train a neural network $\boldsymbol{\theta}$ to predict the policy and value functions at every state following the supervision provided by MCTS.
Specifically, the policy target is the policy generated by MCTS, while the value target is the true discounted return of the state in this episode.

As for state $\tilde{s}^k$ in the MCTS, the policy function provides a prior distribution over actions $\pi_{\boldsymbol{\theta}}^k(\cdot | \tilde{s}^k)$. Actions with high prior probabilities are assigned high pUCT scores, prioritizing their exploration during the search process. However, as the exploration progresses, the influence of this prior gradually diminishes (see details in \cref{sec: MCTS simulation details}). The value function $v_{\boldsymbol{\theta}}^k$ estimates the return and provides the initial value of $\tilde{s}^k$ when $\tilde{s}^k$ is first reached. 

The network $\boldsymbol{\theta}$ is updated based on the overall loss:
\begin{align}
    L(\boldsymbol{\theta})=L_{p}(\pi_{MCTS}, \pi_{\boldsymbol{\theta}}) + L_{v}(r_i, v_{\boldsymbol{\theta}}),
    \label{eq: MARL loss}
\end{align}
where
\begin{align*}
    L_{p}(\pi_1, \pi_2) = \mathbb{E}[-\sum\nolimits_{a \in A_i} \pi_1(a|s^{K,t})\log(\pi_2(a|s^{K,t})],
\end{align*}
\vspace{-1.5em}
\begin{align*}
    L_{v}(r_i,v) = \mathbb{E}[( v(s^{K,t}) - \sum\nolimits_{l=t}^{\infty}\gamma^{l-t} r_i^{K,l} )^2].
\end{align*}

\section{Experiments}

\subsection{Experimental Setup}
\label{setup}
Agents are tested in Markov Stag-Hunt (MSH) and Markov Snowdrift Game (MSG). %

\begin{wrapfigure}[21]{r}{0pt}
    \centering
    \subfigure[MSH]{
    \label{Sequential Stag-hunt game} %
    \includegraphics[width=0.24\linewidth]{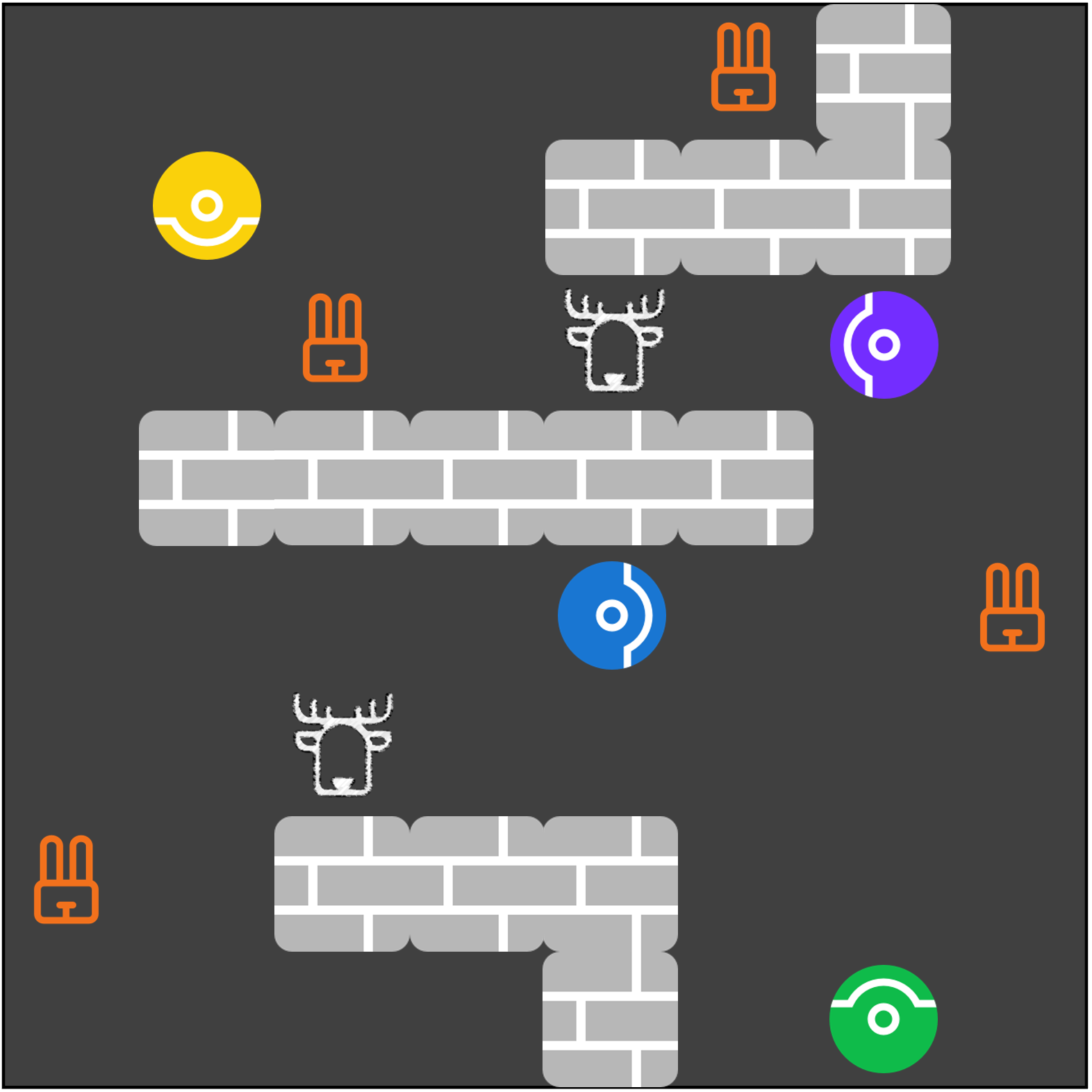}}
  \subfigure[MSG]{
    \label{Sequential Snowdrift game} %
    \includegraphics[width=0.24\linewidth]{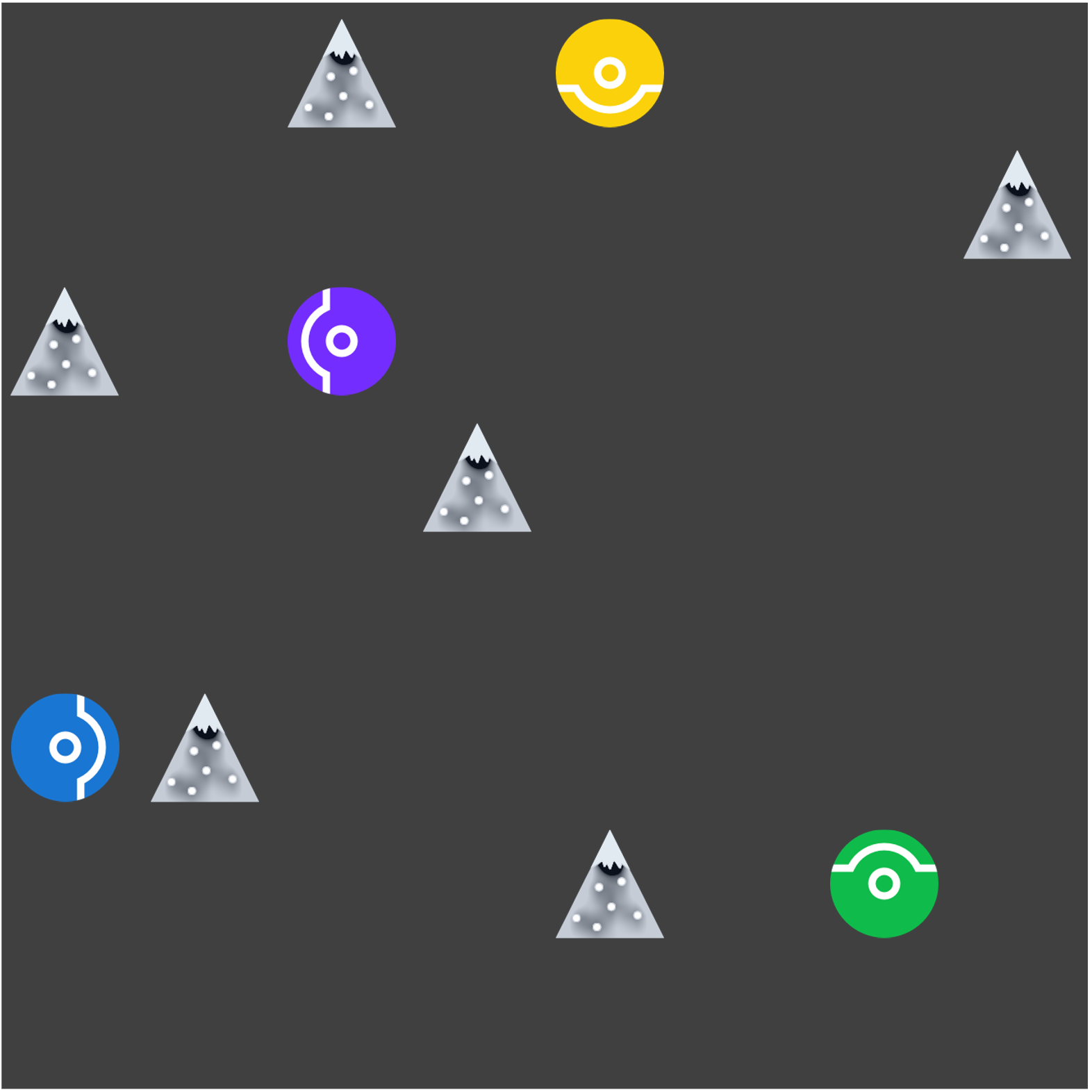}}  
    \caption{Overview of Markov Stag-Hunt and Markov Snowdrift. There are four agents, represented by colored circles, in each paradigm. (a) Agents catch prey for reward. A stag with a reward of $10$ requires at least two agents to hunt together. One agent can hunt a hare with a reward of $1$. (b) Everyone gets a reward of $6$ when an agent removes a snowdrift.  When a snowdrift is removed, removers share the cost of $4$ evenly.}
    \label{symptom_based}
\end{wrapfigure}

\textbf{MSH} expands the environment in \citet{peysakhovich2018prosocial} in terms of the number of agents. In MSH, $4$ agents are rewarded for hunting prey. As shown in \cref{Sequential Stag-hunt game}, each agent has six actions: idle, move left, move right, move up, move down, and hunt. If there are obstacles or boundaries in an agent's moving direction, its position stays unchanged. 
Agents can hunt prey in their current grid.
There are two types of prey: stags and hares. 
A stag provides a reward of $10$, and requires at least two agents located at its grid to execute ``hunt'' together. These cooperating agents will split the reward evenly. 
A hare, which an agent can catch alone, provides a reward of $1$.
After a successful hunt, both the hunters and the prey disappear from the environment.
The game terminates when the timestep reaches $T_{max} = 30$.

We conducted experiments in two different settings of MSH. In the first setting, there are 4 hares and 1 stag (\textbf{MSH-4h1s}). In this scenario, agents can cooperate in hunting the stag to maximize their profits, while also competing with co-players for the opportunity to hunt. The second setting contains 4 hares and 2 stags (\textbf{MSH-4h2s}). There are sufficient stags for agents to cooperate, but the environment will end 5 timesteps after the first successful hunting in each episode. This setup maintains the tension between payoff-dominant cooperation and risk-dominant defection, highlighting the dilemma inherent in the Stag-Hunt game.

In \textbf{MSG} (\cref{Sequential Snowdrift game}), there are six snowdrifts located randomly in an $8\times8$ grid. Similar to MSH, at every time step the agent can stay idle or move one step in any direction. Agents are additionally equipped with a ``remove a snowdrift'' action, which removes the snowdrift in the same cell as the agent. 
When a snowdrift is removed, removers share the cost of $4$ evenly, and every agent gets a reward of $6$.
The game ends when all the snowdrifts are removed or the time $T_{max}=50$ runs out.
The game's essential dilemma arises from the fact that an agent can obtain a higher reward by free-riding, i.e., waiting for co-players to remove the snowdrifts, than by removing a snowdrift themselves.
However, if all agents take free rides, no snowdrift is removed, and agents will not receive any reward.
On the other hand, if any agent is satisfied with a suboptimal strategy and chooses to remove snowdrifts, both the group benefit and individual rewards increase.

In both environments, four agents have no access to each other's parameters, and communication is not allowed. \cref{goal definition} introduces the goal definition of these games. 

\begin{figure}[t]
\centering

  \subfigure[MSH-4h1s]{
    \label{schelling_ssh_4h1s} %
    \includegraphics[width=0.3\linewidth]{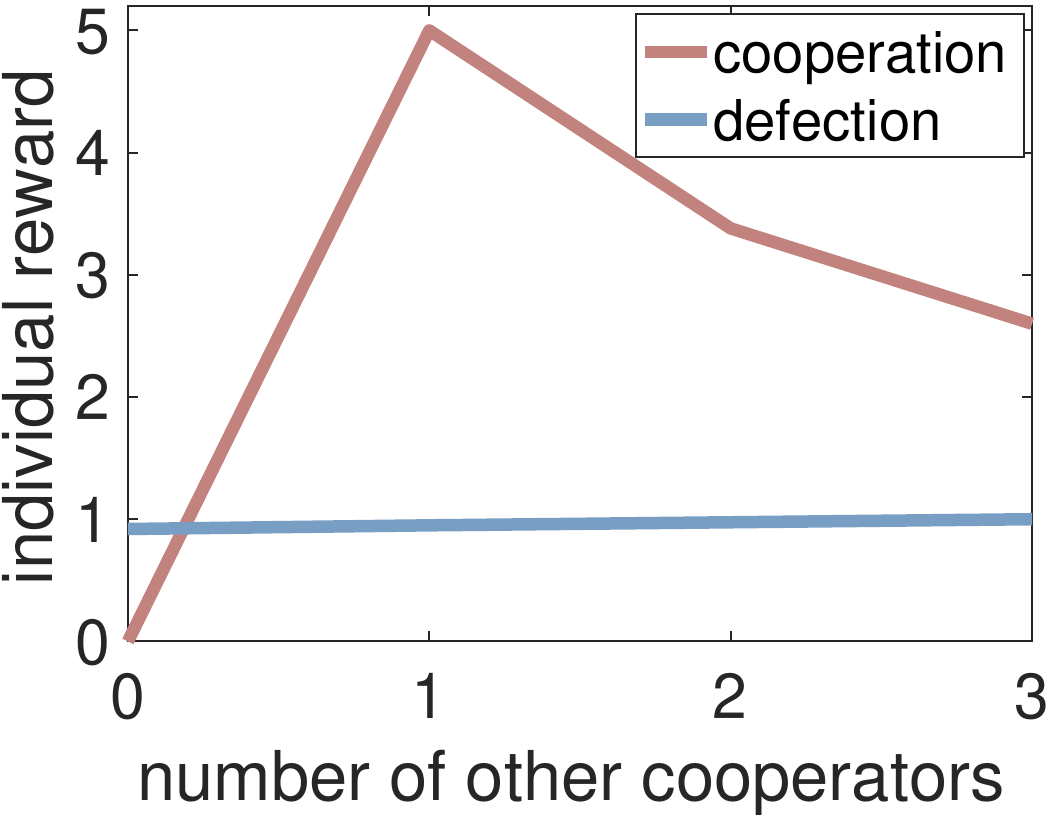}}
  \subfigure[MSH-4h2s]{
    \label{schelling_ssh_4h2s} %
    \includegraphics[width=0.3\linewidth]{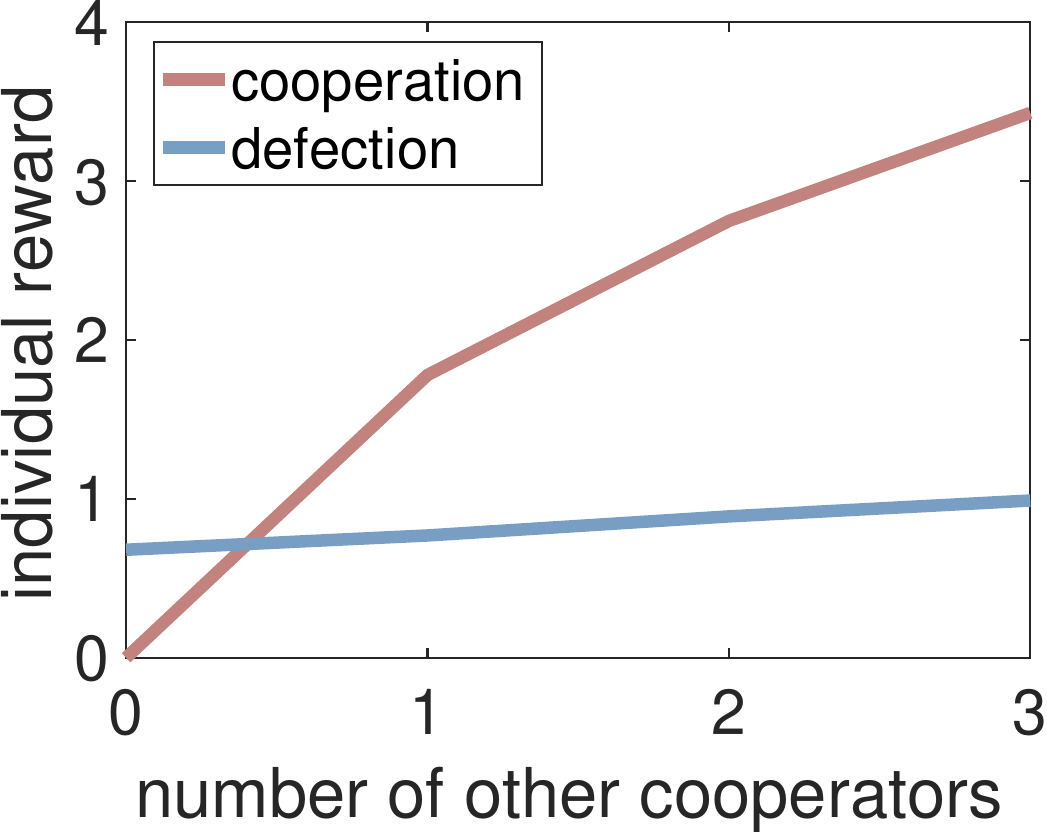}}  
  \subfigure[MSG]{
    \label{schelling_ss} %
    \includegraphics[width=0.3\linewidth]{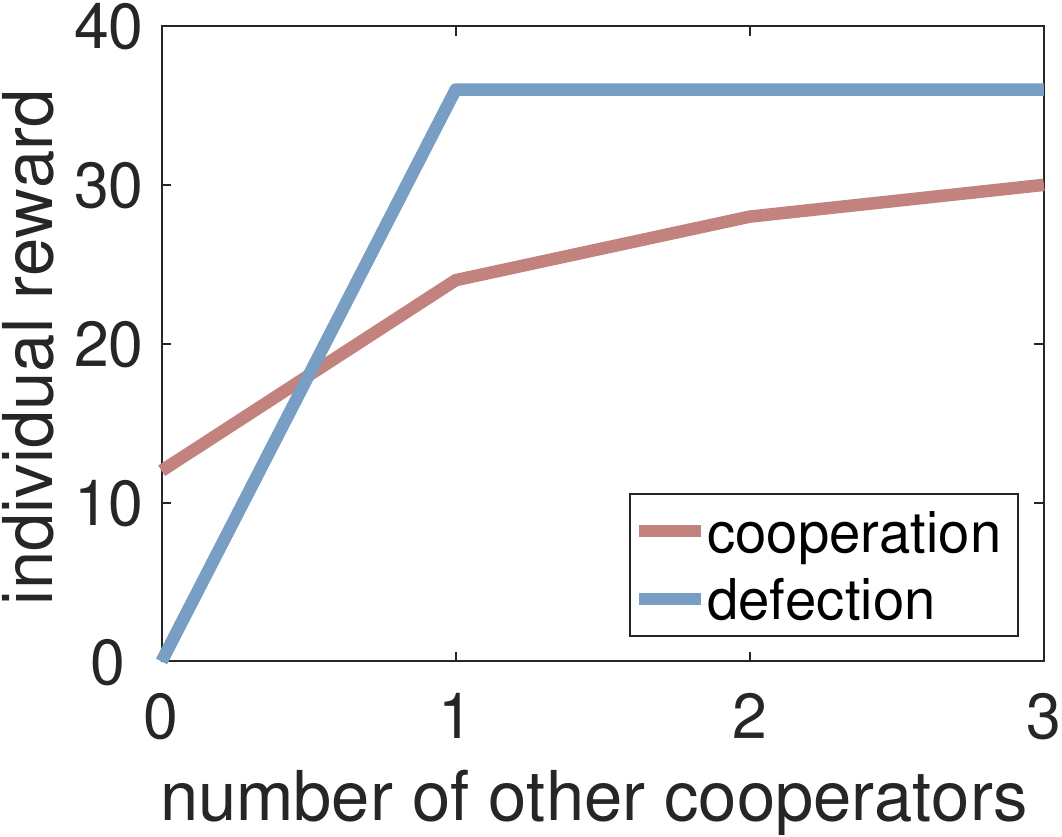}}
    \caption{Schelling diagrams for (a) MSH-4h1s, (b) MSH-4h2s, and (c) MSG. }
    \label{schelling}
     \vspace{-1.5em}
\end{figure}

\vspace{-0.5em}
\paragraph{Schelling diagrams}
Game types are determined by the relative values of elements in the payoff matrix. 
The Schelling diagram~\cite{schelling1973hockey, hughes2018inequity} is a natural generalization of the payoff matrix for two-player games to multi-player settings.
As shown in \cref{schelling}, Schelling diagrams validate our temporal and spatial extension of the matrix-form games, which maintains the dilemmas described by matrix-form games (see a detailed discussion in \cref{Schelling diagram}). 
Moreover, across these three Schelling diagrams, the lines of cooperation and defection intersect. 
This implies that best responses change with co-players' behavior, rendering few-shot adaptation in these environments inherently challenging.

\paragraph{Baselines}
Here, some baseline algorithms are introduced to evaluate the performance of HOP.
During the evaluation of few-shot adaptation, baseline algorithms serve a dual purpose. Firstly, they act as unfamiliar co-players during the evaluation process to test the few-shot adaptation ability of HOP. Secondly, we evaluate the few-shot adaptation ability of the baseline algorithms to demonstrate HOP's superiority.
\emph{LOLA}~\cite{foerster2018learning, zhao2022proximal} agents consider a 1-step look-ahead update of co-players, and update their own policies according to the updated policies of co-players.
\emph{SI}~\cite{jaques2019social} agents have an intrinsic reward term that incentivizes actions maximizing their influence on co-players' actions. The influence is accessed by counterfactual reasoning.
\emph{A3C}~\cite{mnih2016asynchronous} agents are trained using the Asynchronous Advantage Actor-Critic method, a well-established reinforcement learning (RL) technique.
\emph{Prosocial-A3C} (PS-A3C)~\cite{peysakhovich2018prosocial} agents are trained using A3C but share rewards between players during training, so they optimize the per-capita reward instead of the individual reward, emphasizing cooperation between players.
\emph{PR2}~\cite{wen2019probabilistic} agents model how the co-players
would react to their potential behavior, based on which agents find the best response.
The ablated version of HOP, \emph{direct-OM}, retains the planning module, but uses neural networks to model co-players directly (see details in \cref{sec: ablation study}).
In addition, we construct some rule-based strategies.
\emph{Random} policy takes a valid action randomly at each step. 
An agent that consistently adopts cooperative behavior is called \emph{cooperator}, and an agent that consistently adopts exploitative behavior is called \emph{defector}.
In MSH, the goals of cooperators and defectors are hunting the nearest stag and hare, respectively. In MSG, cooperators keep moving to remove the nearest snowdrift, and defectors randomly take actions other than "remove a snowdrift".
When evaluating few-shot adaptation, the set of unfamiliar co-players includes LOLA, A3C, and PS-A3C, serving as representatives of learning agents with explicit opponent modeling module, self-interest purpose, and prosocial purpose, respectively. The co-players also include rule-based agents: random, cooperator and defector.

\subsection{Performance}

\begin{figure}[ht]
\centering
\includegraphics[width=0.7\linewidth]{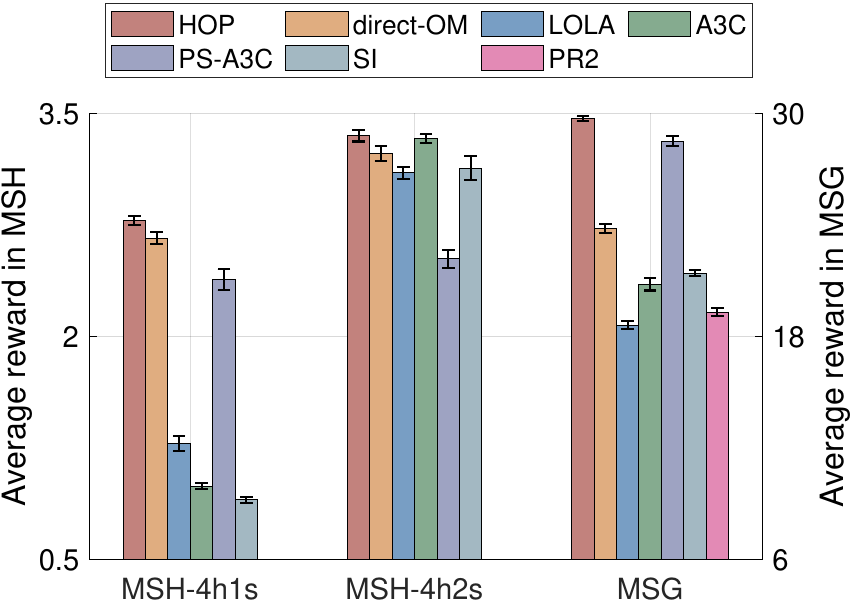}
\caption{Self-play performance of HOP and baseline algorithms. Shown is the average reward in the self-play training phase.}
\label{self-play performance}
\end{figure}

The experiment consists of two phases. 
The first phase focuses on self-play, where agents using the same algorithm are trained until convergence. 
Self-play performance, showing the ability to achieve cooperation, is measured by the algorithm's average reward after convergence.
The second phase evaluates the few-shot adaptation ability of HOP. 
Specifically, a focal agent interacts with three co-players using a different algorithm for 2400 steps. The focal agent's average reward during the final 600 steps is used to measure its algorithm's few-shot adaptation ability. 
At the start of the adaptation phase, any policy's parameters are the convergent parameters derived from the corresponding algorithms in self-play.
During this phase, policies can update their parameters if possible.
Implementation details are given in \cref{sec:implementation detail}.
The results of self-play and few-shot adaptation are displayed in \cref{self-play performance} and \cref{table: adaptation}, respectively.

\paragraph{MSH-4h1s} 
In MSH-4h1s, only HOP, direct-OM, and PS-A3C learn the strategy of hunting stags (\cref{self-play performance}). However, since PS-A3C can get rewards without hunting by itself, it may not effectively learn the relationship between hunting and receiving rewards, leading to a "lazy agent" problem~\cite{sunehag2017value} for PS-A3C. This results in the overall reward of PS-A3C being inferior to HOP and direct-OM. LOLA swings between hunting stags and hunting hares. SI and A3C primarily learn the strategy of hunting hares, resulting in low rewards. PR2 fails to work in MSH. In this environment, the number of agents may be reduced due to the successful hunting of agents, and this is not supported by PR2. Despite attempts to modify the algorithm accordingly, the modified version ultimately failed to learn a decent policy. As a result, the relevant results of PR2 in MSH are not shown in \cref{self-play performance} and \cref{table: adaptation}.

HOP learns the stag hunting strategy through self-play, enabling seamless cooperation with agents like PS-A3C and cooperators, which similarly prioritize stag hunting (\cref{subtable: SH_4h1s}). This compatibility stems from the fact that in the Stag-Hunt game, the best response of cooperation is cooperation. Thus, direct-OM and PS-A3C agents, who are equipped with learned cooperative strategies, also attain relatively high rewards when playing with cooperative co-players. 
When confronted with co-players with fluctuating strategies such as LOLA or random agents lacking fixed objectives, HOP seeks out opportunities for heightened returns through cooperation. 
Furthermore, when encountering co-players like A3C and defectors, known for their inclination towards hunting hares, HOP adjusts to these non-cooperative scenarios within a small amount of interaction. HOP and direct-OM achieve substantially greater rewards when confronting defectors compared to PS-A3C, who also favors cooperation. This observation highlights the pivotal role of the planning module in efficient adaptation.

\paragraph{MSH-4h2s}
As depicted in \cref{self-play performance}, in MSH-4h2s, all algorithms have learned the strategy of cooperatively hunting stags, among which HOP and A3C are more stable and yield higher returns. PS-A3C tends to delay hunting, as early hunting results in leaving the environment and failing to obtain the group reward from subsequent hunting. This may lead PS-A3C to suboptimal actions in the last few steps and thus fail to hunt under the 5-step termination rule.

The adaptation performance in MSH-4h2s is presented in \cref{subtable: SH_table_modified}. When facing with the cooperator, the best response is to hunt stags, which requires minimal adjustments to each algorithm's policies, so their returns are comparable to the Orcale reward. Similarly, when encountering learning co-players who have adopted the cooperation policy, HOP and most baselines yield high rewards. However, given that learning agents may dynamically adjust their goals, it becomes essential to discern the real-time goals of the co-players in order to find the best response. In these scenarios, HOP's performance surpasses that of other algorithms, approaching the Orcale reward. When playing with non-cooperative co-players such as random and defectors, significant strategy adjustments are necessary for each algorithm to achieve high returns. Therefore, the returns for all algorithms are notably diminished. HOP demonstrates superior adaptability compared to the other algorithms, exhibiting its ability to make substantial strategic adjustments.

\begin{figure}[t]
    \centering
    \subfigure[MSH-4h1s]{
    \includegraphics[width=0.45\linewidth]{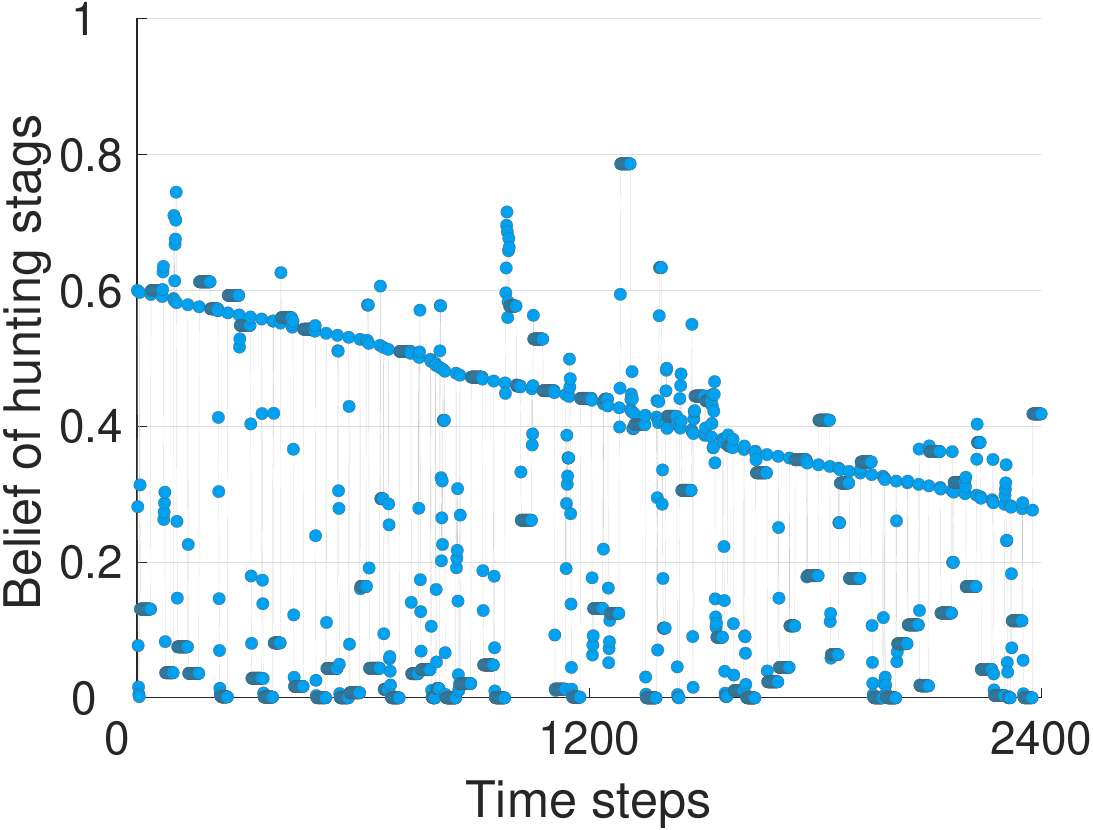}
    }
    \subfigure[MSH-4h2s]{
    \includegraphics[width=0.45\linewidth]{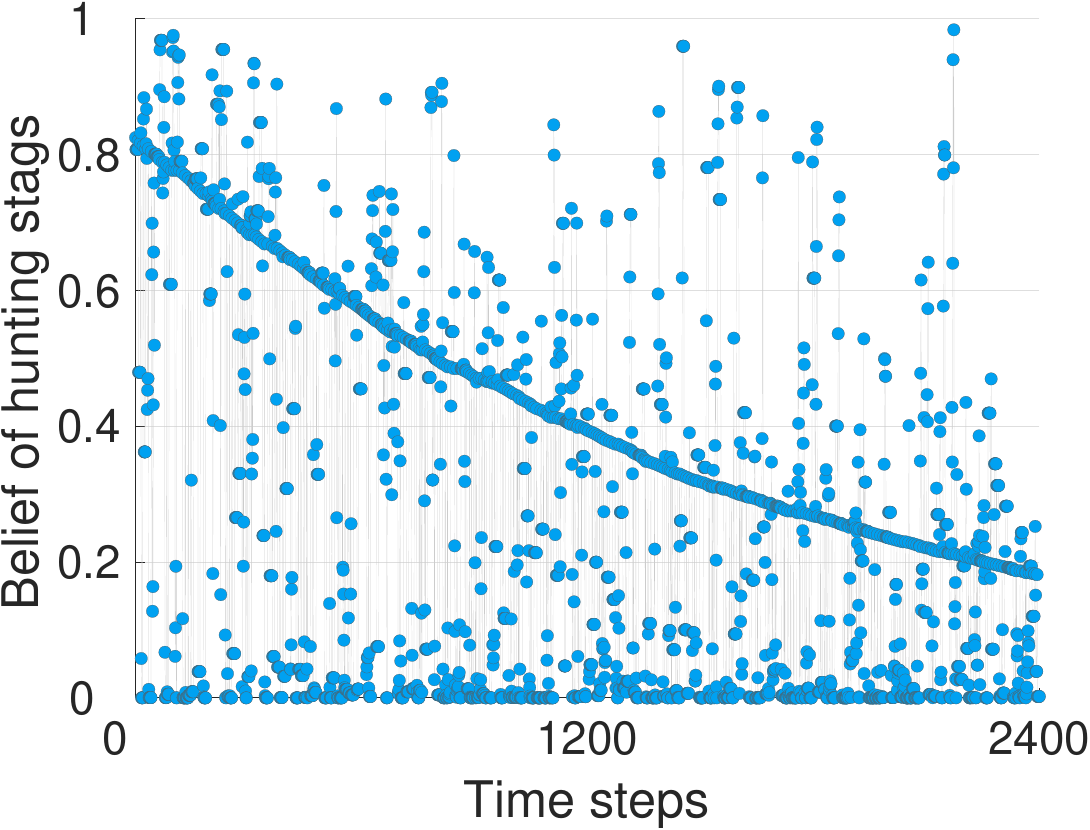}
    }
    \caption{Visualization of HOP's belief in adaptation to three defectors in MSH. Every blue-filled circle represents HOP's inferred probability (i.e., belief) that a co-player hunts stags.}
    \label{visua-beliefs}
\end{figure}

\begin{table*}[ht]

\caption{Few-shot adaptation performance of HOP and baselines in (a) MSH-4h1s, (b) MSH-4h2s, and (c) MSG. The interaction happens between 1 agent using the row policy and 3 co-players using the column policy. Shown are the min-max normalized rewards, with normalization bounds set by the rewards of Orcale and the lowest rewards among all baselines and random policy.
See detailed description and analysis of Orcale in \cref{sec: long interaction}. The results are depicted for the row policy from 1800 to 2400 step. Overall best adaptation percentage shows the proportion of scenarios in which the algorithm performs optimally, while accounting for standard deviation. }
\centering
\label{table: adaptation}

\subtable[Performance in MSH-4h1s]{
\renewcommand\arraystretch { 0.8}
\centering
\begin{tabular}{c c c c c c c c}
    \hline
        & \multicolumn{3}{c}{learning co-players} & \multicolumn{3}{c}{rule-based co-players} & \multirow{2}{*}{Overall best}\\ \cmidrule(lr){2-4} \cmidrule(lr){5-7}
        & LOLA &A3C & PS-A3C & random & cooperator & defector & adaptation percentage\\ \hline
        HOP & \textbf{0.97}\err{0.06} &\textbf{0.80}\err{0.15} & \textbf{0.93}\err{0.04} & \textbf{0.96}\err{0.07} & \textbf{0.74}\err{0.06} & 0.51\err{0.02} & \textbf{83.3\%}\\
        direct-OM & 0.64\err{0.05} & 0.57\err{0.07} & 0.79\err{0.04} & \textbf{0.91}\err{0.10} & 0.56\err{0.04} & 0.44\err{0.04} & 16.7\%\\
        LOLA & - & 0.55\err{0.07} & 0.38\err{0.04} & 0.45\err{0.06} & 0.46\err{0.03} & 0.41\err{0.02} & 0.0\%\\
        A3C & 0.35\err{0.01}  & - & 0.24\err{0.02} & 0.85\err{0.01} & 0.31\err{0.02} & \textbf{1.00}\err{0.01} & 20.0\%\\ 
        PS-A3C & 0.85\err{0.05}  & 0.60\err{0.11} & - & 0.64\err{0.06} & 0.55\err{0.05} & 0.09\err{0.04} & 0.0\%\\
        SI &  0.32\err{0.01} & \textbf{0.95}\err{0.04} & 0.22\err{0.02}  &  0.81\err{0.01} & 0.28\err{0.02} & 0.89\err{0.04} & 16.7\% \\
        
        \hline
    \end{tabular}
    \label{subtable: SH_4h1s}
}

\subtable[Performance in MSH-4h2s]{
\renewcommand\arraystretch { 0.8}
\centering
\begin{tabular}{c c c c c c c c}
    \hline
        & \multicolumn{3}{c}{learning co-players} & \multicolumn{3}{c}{rule-based co-players} & \multirow{2}{*}{Overall best} \\ \cmidrule(lr){2-4} \cmidrule(lr){5-7}
        & LOLA &A3C & PS-A3C & random & cooperator & defector & adaptation percentage\\ \hline
        HOP & \textbf{0.97}\err{0.02}  &\textbf{0.99}\err{0.02} & \textbf{0.88}\err{0.02} & \textbf{0.78}\err{0.07} & \textbf{1.00}\err{0.01} & \textbf{0.36}\err{0.02} &  \textbf{100.0\%}\\
        direct-OM & \textbf{0.95}\err{0.01} &0.85\err{0.02} & 0.74\err{0.03} & 0.62\err{0.04} & 0.96\err{0.02} & 0.31\err{0.02} & 16.7\%\\
        LOLA & - & 0.92\err{0.04} & 0.82\err{0.02} & \textbf{0.75}\err{0.04} & \textbf{1.00}\err{0.03} & 0.28\err{0.03}  & 40.0\%\\
        A3C &  0.91\err{0.02} & - & \textbf{0.87}\err{0.02} & 0.55\err{0.05} & \textbf{0.98}\err{0.02} & 0.25\err{0.02}  & 40.0\%\\ 
        PS-A3C  & 0.24\err{0.03} &0.18\err{0.02} & - & 0.29\err{0.02} & 0.38\err{0.01} & 0.06\err{0.02} & 0.0\% \\
        SI & 0.77\err{0.02} & 0.83\err{0.01} & 0.74\err{0.01} & 0.52\err{0.03} & 0.87\err{0.03} & 0.27\err{0.02} & 0.0\% \\
    \hline
    \end{tabular}
\label{subtable: SH_table_modified}
}
\subtable[Performance in MSG]{
\renewcommand\arraystretch { 0.8}
\centering
\begin{tabular}{c c c c c c c c}
    \hline
        & \multicolumn{3}{c}{learning co-players} & \multicolumn{3}{c}{rule-based co-players} & \multirow{2}{*}{Overall best} \\ \cmidrule(lr){2-4} \cmidrule(lr){5-7}
        & LOLA & A3C & PS-A3C & random & cooperator & defector & adaptation percentage\\ \hline
        HOP & \textbf{0.78}\err{0.04}  &\textbf{0.39}\err{0.09} & \textbf{0.65}\err{0.08} & \textbf{0.44}\err{0.03} & 0.48\err{0.05} & \textbf{0.55}\err{0.01} & \textbf{83.3\%}\\
        direct-OM & 0.31\err{0.11} & 0.12\err{0.05} & \textbf{0.55}\err{0.04}& \textbf{0.38}\err{0.04}& \textbf{0.67}\err{0.05} & 0.34\err{0.05} & 50.0\%\\ 
        LOLA & - & \textbf{0.33}\err{0.07}& \textbf{0.55}\err{0.06}& 0.25\err{0.08} & 0.43\err{0.04} & 0.18\err{0.01} & 40.0\%\\
        A3C & 0.33\err{0.04} & - & \textbf{0.52}\err{0.09}& 0.30\err{0.04} & 0.33\err{0.03} & 0.14\err{0.01} & 20.0\%\\ 
        PS-A3C & 0.67\err{0.05} & \textbf{0.35}\err{0.04} & - & 0.33\err{0.04} & 0.00\err{0.08} & 0.38\err{0.02} & 20.0\%\\ 
        SI & \textbf{0.74}\err{0.08} & 0.00\err{0.05} & 0.33\err{0.08} & 0.00\err{0.04} & 0.24\err{0.07} & 0.24\err{0.03} & 16.7\%\\
        PR2 & 0.00\err{0.13} & 0.00\err{0.08} & \textbf{0.58}\err{0.05} & 0.16\err{0.05} & 0.43\err{0.02} & 0.14\err{0.01} & 16.7\%\\
        \hline
    \end{tabular}
\label{subtable: SD_table_modified}
}
\end{table*}

We would like to provide further intuition on why HOP is capable of efficiently adapting its policy to unseen agents. Take the experiment facing three defectors (always attempting to hunt the nearest hare) as an example. There are two goals here: hunting stags or hunting hares. At the start of the evaluation phase, HOP holds the belief that every co-player is more likely to hunt a stag because HOP has seen its co-players hunt stags more than hares during self-play. This false belief for defectors degrades HOP's performance. Both intra-OM and inter-OM correct this false belief by updating during the intereactions with defectors (see visualization of belief update in \cref{visua-beliefs}). Intra-OM provides the ability to correct the belief of hunting stags within an episode. Specifically, as a co-player keeps moving closer to a hare, intra-OM will update the belief of the co-player toward the goal ``hare'', leading to accurate opponent models. In \cref{visua-beliefs}, there are many points with values near 0, showing that HOP infers that the agent's goal is unlikely to be a stag through intra-OM. Taking these accurate co-player policies as input, the planning module can output advantageous actions. Inter-OM further accelerates the convergence towards true belief by updating the inter-episode belief, which is used as a prior for intra-OM at the start of every episode. A declining line, formed by the points from initial steps of each episode, appears in both sub-figures of \cref{visua-beliefs}, which reflects that HOP gradually reduces the prior of the co-player hunting a stag through inter-OM.

\paragraph{MSG} As shown in \cref{self-play performance}, HOP achieves the highest reward during self-play and it is close to the theoretically optimal average reward in this environment (i.e. when all snowdrifts are removed, resulting in a group average reward of 30.0). This outcome is a remarkable achievement in a fully decentralized learning setting and highlights the high propensity of HOP to cooperate. In contrast, LOLA, A3C, SI, and PR2 prioritize maximizing their individual profits, which leads to inferior outcomes due to their failure to coordinate and cooperate effectively. PS-A3C performs exceptionally well in self-play, ranking second only to HOP. Like in MSH, it fails to achieve the maximum average reward due to the coordination problem, which is prominent when only one snowdrift is left. This issue highlights the instability of the policy due to the absence of action planning.

HOP demonstrates the most effective few-shot adaptation performance (\cref{subtable: SD_table_modified}). Specifically, when adapting to three defectors, HOP receives substantially higher rewards than other policies. This highlights the effectiveness of HOP in quickly adapting to non-cooperative behavior, which differs entirely from behavior of co-players in HOP's self-play. In contrast, A3C and PS-A3C do not explicitly consider co-players. They have learned the strategies tending to exploit and cooperate, respectively. Therefore, A3C performs effectively against agents that have a higher tendency to cooperate, such as the cooperator. However, its performance is relatively poor when facing non-cooperative agents. Conversely, PS-A3C exhibits the opposite behavior. %

Overall, the above experiments demonstrate the remarkable adaptation ability of HOP across all environments (see last columns in \autoref{table: adaptation}). Other algorithms can only achieve the best adaptation performance when facing some specific co-players, to whom the best response is close to the policies learned by the algorithms in self-play. HOP can achieve the best adaptation level in most test scenarios, where co-players perform either familiar or completely unfamiliar behavior. Meanwhile, HOP exhibits advantages during self-play.

Ablation study indicates that inter-OM and intra-OM play crucial roles in adapting to agents with fixed goals and agents with dynamic goals, respectively. Moreover, if opponent modeling is not conditioned on goals, the self-play and few-shot adaptation abilities are greatly weakened. Further details are provided in \cref{sec: ablation study}. 

We observe the emergence of social intelligence, including self-organized cooperation and an alliance of the disadvantaged, during the interaction of multiple HOP agents in mixed-motive environments. Further details can be found in \cref{sec:emergence}.

\section{Conclusion and Discussion}
We propose Hierarchical Opponent modeling and Planning (HOP), a hierarchical algorithm for few-shot adaptation to unseen co-players in mixed-motive environments. It consists of an opponent modeling module for inferring co-players' goals and behavior and a planning module guided by the inferred information to output the focal agent's best response. Empirical results  show that HOP performs better than state-of-the-art MARL algorithms, in terms of dealing with mixed-motive environments in the self-play setting and few-shot adaptation to previously unseen co-players.

Whilst HOP exhibits superior abilities, there are several limitations illumining our future work. First, in any environment, a clear definition of goals is needed for HOP. To enhance HOP's ability to generalize to various environments, a technique that can autonomously abstract goal sets in various scenarios is needed, which \cite{ashwood2022dynamic} has attempted to explore. Second, we use Level-0 ToM, which involves "think of what they think." However, a more complex form of ToM, such as Level-1 ToM that considers "what I think they think about me," has the potential to improve our predictions about co-players. Nevertheless, incorporating nested inference introduces a higher computational cost. Consequently, it becomes imperative to develop advanced planning methods that can effectively and rapidly leverage the insights provided by high-order ToM. Third, we investigate mix-motive environments with the expectation that HOP can facilitate effective decision-making and adaptation in human society. Despite selecting diverse well-established algorithms as co-players, none of them adequately model human behavior. It would be interesting to explore how HOP can perform in a few-shot adaptation scenario involving human participants. As HOP is self-interested, it may not always align with the best interest of humans. One way to mitigate this risk is leveraging HOP's ability to infer and optimize for human values and preferences during interactions, thereby assisting humans in complex environments.

\section*{Acknowledgements}
This project is supported by the National Key R\&D Program of China (2022ZD0114900).

\section*{Impact Statement}
This paper presents work whose goal is to advance the field of Machine Learning. There are many potential societal consequences of our work, none which we feel must be specifically highlighted here.

\bibliography{our_paper}
\bibliographystyle{icml2024}

\newpage
\appendix
\onecolumn
\section{Pseudo Code of HOP}
\label{sec: pseudo code}

\begin{algorithm}
    \caption{HOP}
    \label{alg: HOP}
    \begin{algorithmic}
    \STATE {\bfseries Input:} Number of MCTS tree $N_s$, update interval $T_u$, capacity of the trajectory buffer $L$, goal set $G_j$ $(j \neq i)$, initial belief of agents' goals $b_{ij}^{0,0}(g_j)$.
    \STATE {\bfseries Output:} Actions $a_i^{K,t}$, planning module network $\boldsymbol{\theta}$, goal-conditioned policy network $\boldsymbol{\omega}$.
        \FOR {each episode $K$}
            \STATE generate initial state of this episode $s^{K, 0}$ randomly
            \FOR {$t = 0$ {\bfseries to} $T_{max} - 1$}
                \REPEAT
                    \STATE sample $\mathbf{g}_{-i}^l$ from $b_{ij}^{K,t}(g_j) (j \neq i)$
                    \STATE get $Q_l(s^{K,t}, a, \mathbf{g}_{-i}^l)$ $(\forall a)$ via MCTS
                \UNTIL {$N_s$ times}
                \STATE calculate $Q_{avg}(s^{K,t}, a)$ $(\forall a)$ [\cref{eq: Qavg}]
                \STATE choose action $a_i^{K,t}$ from $\pi_{MCTS}(a|s^{K,t})$ [\cref{eq: pi_MCTS}]
                \STATE intra-OM update $b_{ij}^{K,t+1}$ [\cref{eq: intra}]
                \STATE collect data of this step to the trajectory buffer
            \ENDFOR
            \IF {the trajectory buffer is full}
                \STATE update $\boldsymbol{\omega}$ [\cref{eq: modeling loss}]
            \ENDIF
            \IF {$K \times T_{max} \equiv 0$ (mod $T_u$) }
                \STATE update $\boldsymbol{\theta}$ [\cref{eq: MARL loss}]
            \ENDIF
            \STATE inter-OM update $b_{ij}^{K+1,0}$ [\cref{eq: inter}]
        \ENDFOR
    \end{algorithmic}
\end{algorithm}

\section{Theoretical Analysis}
\label{sec:theoretical analysis}
We aim to offer a concise theoretical analysis. Due to the complexity of environments characterized by both temporal and spatial structures, attaining theoretical guarantees in such environments can be inherently challenging. To strike a balance, we have undertaken a verification of the theoretical guarantee associated with HOP in the matrix games. These games encapsulate the same dilemma of sequential games.  For clarity, our analysis will be conducted in the context of a two-player game, and the analysis can be extended to games involving a greater number of agents.
Consider a two-player game where both players have two goals: ``Cooperate" and ``Defect," resulting in a utility matrix shown in \cref{table:utility-matrix}.

\begin{table}[ht]
\centering
\caption{Utility matrix for a two-player game. Each element in the table represents the utility of the row player (first value) and the utility of the column player (second value). The utility values $R$, $S$, $T$, and $P$ determine different game paradigms.}
\begin{tabular}{c|cc}
       & Cooperate & Defect \\ 
\hline
   Cooperate & $R, R$ & $S, T$ \\
   Defect & $T, S$ & $P, P$ \\
\end{tabular}
\label{table:utility-matrix}
\end{table}

Suppose HOP is the row player. At a certain timestep, the column player selects its goal $g_{column}$ to be ``Cooperate" with a probability of $p$ and to be ``efect" with a probability of $1-p$. We sample the co-player's goal to simulate using Monte Carlo Tree Search (MCTS), with a frequency of $p + \epsilon$ to ``Cooperate" and a frequency of $1-p-\epsilon$ to ``Defect."

In the current state $s$, we have two possible actions: $a_1$ for cooperation and $a_2$ for defection. During the MCTS planning process, when the co-player aims to ``Cooperate," we have:
\[Q(s, a_1|g_{\text{column}} = \text{``Cooperate"}) = R(1 + \epsilon_R)\]
\[Q(s, a_2|g_{\text{column}} = \text{``Cooperate"}) = T (1 + \epsilon_T)\]

When the co-player aims to ``Defect," we have:
\[Q(s, a_1|g_{\text{column}} = \text{``Defect"}) = S (1 + \epsilon_S)\]
\[Q(s, a_2|g_{\text{column}} = \text{``Defect"}) = P (1 + \epsilon_P)\]

Thus, we can calculate the overall Q-values as follows:
\[Q(s, a_1) = (p+\epsilon)R(1 +\epsilon_R) + (1 - p -\epsilon)S(1+\epsilon_S)\]
\[Q(s, a_2) = (p+\epsilon)T(1+\epsilon_T)+(1 -p - \epsilon)P(1+\epsilon_P)\]

In the learning process, the goal-conditioned policy network is trained using supervised learning, and its accuracy significantly improves with sufficient rounds of observation. Consequently, the accuracy of the environment simulation within the Monte Carlo Tree Search (MCTS) algorithm becomes exceedingly high. In such a scenario, the convergence guarantee of MCTS remains intact, resulting in a final precision of MCTS that is remarkably high. Specifically, we have $|\epsilon_R|, |\epsilon_S|, |\epsilon_T|, |\epsilon_P| \ll |\epsilon|$, and these small error terms can be safely ignored.

Then, when 
\[\frac{T+S-R-P}{p(R-T)+(1-p)(S-P)} \epsilon < 1,\]
the optimal strategy that HOP obtains is consistent with the true optimal strategy. Two factors affect the size of $|\epsilon|$: the accuracy in inferring the co-player's goals and the deviation between frequency and probability when sampling the goal. To address the accuracy issue, we employ two layers of modules, intra-OM and inter-OM, to make accurate predictions as early as possible in each episode. For the deviation between frequency and probability, we increase the value of $N_s$ to reduce this deviation. In practical applications, the choice of an appropriate $N_s$ depends on the trade-off between computational speed and sampling accuracy.

\section{Goal Definition}
\label{goal definition}
In MSH, we define two goals: $g^C$ as hunting stags and $g^D$ as hunting hares.

In MSG, we define two goals: $g^C$ as removing the drifts, and $g^D$ as staying lazy (i.e. not attempting to remove any snowdrifts). For inter-OM, the goal $g^C$ is decomposed into 6 parts: $g^{Ck}\ (1 \leq k \leq 6)$, where $g^{Ck}$ represents removing $k$ snowdrift(s) in one episode. $b_{ij}^{K,0}(g^{Ck})$ and $b_{ij}^{K,0}(g^D)$ will be updated according to \cref{eq: inter}. During an episode, if the co-player $j$ has removed $m$ snowdrift(s) at time $t$ of the episode $K$, our belief $b_{ij}^{K,t}(g_j^C) = \sum_{k=m+1}^6 b_{ij}^{K,0}(g_j^{Ck})$.

For intra-OM, each snowdrift $s$ is defined as a subgoal $g^{C[s]}$. 
We use \cref{eq: intra} conditioned on $g^C$ to update our belief:
\begin{align*}
    \begin{split}
        b_{ij}^{K,t+1}(g_j^{C[s]}|g_j^{C}) &= \frac{1}{Z_1}b_{ij}^{K,t}(g_j^{C[s]}|g_j^{C}) Pr_i(a_j^{K,t} | s^{K,0:t}, g_j^{C[s]}),
    \end{split}
\end{align*}
where $Z_1$ is the normalization factor. 
We can update our belief of an agent removing a snowdrift $s$:
\begin{align*}
    b_{ij}^{K,t}(g^{C[s]})=b_{ij}^{K,t}(g_j^{C[s]}|g_j^{C}) b_{ij}^{K,t}(g_j^C).
\end{align*}
At the start of an episode, $b_{ij}^{K,0}(g_j^{C[s]}|g_j^{C})$ is set to be uniform, which means $b_{ij}^{K,0}(g_j^{C[s]}|g_j^{C}) = \frac{1}{6}$. We train the goal-conditioned policy network $\boldsymbol{\omega}$ conditioned on $g^{C[s]}$.

\section{Schelling Diagram}
\label{Schelling diagram}
The Schelling diagram compares the rewards of different potential strategies (i.e., cooperation and defection here) given a fixed number of other cooperators.
It is a natural generalization of the payoff matrix for two-player games to multi-player settings.
Here, we use Schelling diagrams to validate our temporal and spatial extension of the matrix-form games.

\cref{schelling_ssh_4h1s} and \cref{schelling_ssh_4h2s} show the Schelling diagrams of MSH.
Defection (i.e., hunting hare) is a safe strategy as a reasonable reward is guaranteed independent of the co-players’ strategies.
Cooperation (i.e., hunting stag) poses the risk of being left with nothing (when there are no others hunting stag), but is more rewarding if at least one co-player hunts stag. 
That is to say, hunting hare is risk dominant, and hunting stag is reward dominant.
This is consistent with the dilemma described by the matrix-form stag-hunt game~\cite{bloembergen2011lenient}.
In the ``4h1s" setting, when there are more than two cooperators, the choice to act as a cooperator carries the risk of not being able to successfully hunt. In the ``4h2s" setting, the income of cooperators increases with the number of cooperators, resulting in a lower risk of choosing to hunt stag compared to the ``4h1s" setting.

In the matrix-form snowdrift game, cooperation incurs a cost to the cooperator and accrues benefits to both players regardless of whether they cooperate or not ~\cite{souza2009evolution}. 
There are two pure-strategy Nash equilibria: player 1 cooperates and player 2 defects; player 1 defects and player 2 cooperates. 
That is, the best response is playing the opposite strategy from what the coplayer adopts. 
As shown in \cref{schelling_ss}, in MSG, one agent's optimal strategy is cooperation (i.e., removing snowdrifts) when no co-players cooperate, but when there are other cooperators, the optimal strategy is defection (i.e., free-riding).
Our MSG is an appropriate extension of the matrix-form snowdrift game.

\section{Implementation Details}
\label{sec:implementation detail}

\subsection{MCTS Simulation Details}
\label{sec: MCTS simulation details}
As introduced in \cref{sec:planning}, we run MCTS for $N_s$ rounds. In each round, we run $N_i$ search iterations (see \citet{browne2012survey} for details of each iteration). The score of an action $a$ at state $\tilde{s}^k$ is:
\begin{align*}
    Score(\tilde{s}^k, a) = Q(\tilde{s}^k, a) + c \pi_{\boldsymbol{\theta}}(a | \tilde{s}^k) \frac{ \sqrt{\sum\nolimits_{a'}N(\tilde{s}^k, a')} }{1 + N(\tilde{s}^k, a)}
\end{align*}
where $Q(\tilde{s}^k, a)$ denotes the average return obtained by selecting action $a$ at state $\tilde{s}^k$ in the previous search iterations. $N(\tilde{s}^k, a)$ represents the number of times action $a$ has been selected at state $\tilde{s}^k$ in the previous search iterations. $\pi_{\boldsymbol{\theta}}(a | \tilde{s}^k)$ refers to the policy provided by the network $\boldsymbol{\theta}$. $c$ is the exploration coefficient. We select the action which has the highest score when reaching $\tilde{s}^k$ at the selection phase of one search iteration.

\subsection{Network Architecture}
The goal-conditioned policy network $\boldsymbol{\omega}$ and the policy-value network for MCTS $\boldsymbol{\theta}$ both start with three convolutional layers with the kernel size 3 and the stride size 1. Three layers have 16, 32, and 32 output channels, respectively. They are connected to two fully connected layers. The first layer has an output of size 512, and the second layer gives the final output.

\subsection{Hyperparameters}
For each result in \cref{self-play performance}, \cref{table: adaptation}, \cref{table: LI-Ref} and \cref{table: ablation}, we performed 10 independent experiments using different random seeds.
The left-hand side of $\pm$ represents the average reward of the 10 trials, and the right-hand side represents the standard error.

Hyperparameters for HOP are listed in \cref{table: hyperparameter-HOP}. $\alpha$ and $T_u$ are tuned in the adaptation phase to achieve fast adaptation. As $\alpha$ decreases, agents attach greater importance to recent episodes, which will speed up the adaptation to new behaviors of the co-players. It is not advisable to adjust $\alpha$ too small, otherwise the update may be unstable due to the randomness of the co-player's strategy.

Hyperparameters for baselines are listed in \cref{table: hyperparameter}. Some hyperparameters are tuned in the adaptation phase to achieve fast adaptation.

\begin{table}[ht]
\centering
\caption{Hyperparameters}
\label{table: hyperparameter}
\subtable[HOP]{
\renewcommand\arraystretch { 1.0 }
\centering
\begin{tabular}{c c c c c}
\hline
    
& \multicolumn{2}{c}{self-play phase} & \multicolumn{2}{c}{adaptation phase} \\ \cmidrule(lr){2-3} \cmidrule(lr){4-5}
       & MSH & MSG  & MSH& MSG\\ 
\hline

    horizon weight $\alpha$ & 0.99 & 0.99  & 0.95 & 0.95\\
    rationality coefficient $\beta$ & 2 & 2  & 5 & 5\\
    discount factor $\gamma$ & 0.95 & 0.95  & 0.95& 0.95\\
    update interval $T_u$ & 2000 & 2000  & 200 & 200\\
    capacity of the trajectory buffer $L$ & 5000 & 5000 & 5000& 5000\\
    number of MCTS rounds $N_s$ & 8 & 5  & 8 & 5\\
    number of search iterations for each MCTS $N_i$ & 200 & 200  & 200 & 200\\
    exploration coefficient $c$ & 2 & 12 & 2 & 12\\
    learning rate & $10^{-4}$ & $10^{-4}$  & $5 \times 10^{-4}$ & $5 \times 10^{-4}$\\
    OM learning rate & $5 \times 10^{-4}$ & $5 \times 10^{-4}$  & $5 \times 10^{-4}$ & $5 \times 10^{-4}$\\
    
\hline
\label{table: hyperparameter-HOP}
\end{tabular}
}

\subtable[A3C and PS-A3C]{
\renewcommand\arraystretch { 1.0 }
\centering
\begin{tabular}{c c c c c}
\hline
    
& \multicolumn{2}{c}{self-play phase} & \multicolumn{2}{c}{adaptation phase} \\ \cmidrule(lr){2-3} \cmidrule(lr){4-5}
       & MSH & MSG  & MSH& MSG\\ 
\hline
    learning rate& $10^{-4}$ & $10^{-4}$  & $5 \times 10^{-4}$ & $5 \times 10^{-4}$\\
    batch size & 2000 &  2000  & 200 & 200\\
    discount factor& 0.99 & 0.99  & 0.99 & 0.99\\
    value function loss coefficient& 0.5 & 0.5  & 0.5& 0.5\\
    gradient clip & 40 & 40  &40 &40\\
    entropy coefficient & 0.01 & 0.01 & 0.01& 0.01\\
\hline
\end{tabular}
}

\subtable[LOLA]{
\renewcommand\arraystretch { 1.0 }
\centering
\begin{tabular}{c c c c c}
\hline
    
& \multicolumn{2}{c}{self-play phase} & \multicolumn{2}{c}{adaptation phase} \\ \cmidrule(lr){2-3} \cmidrule(lr){4-5}
       & MSH & MSG  & MSH& MSG\\ 
\hline
    learning rate & $10^{-4}$ & $10^{-4}$  & $5 \times 10^{-4}$ & $5 \times 10^{-4}$\\
    OM learning rate & $5 \times 10^{-4}$ & $5 \times 10^{-4}$  & $5 \times 10^{-4}$ & $5 \times 10^{-4}$\\
    batch size & 2000 &  2000  & 200 & 200\\
    discount factor& 0.99 & 0.99  & 0.99 & 0.99\\
\hline
\end{tabular}
}

\subtable[Social Influence]{
\renewcommand\arraystretch { 1.0 }
\centering
\begin{tabular}{c c c c c}
\hline 
& \multicolumn{2}{c}{self-play phase} & \multicolumn{2}{c}{adaptation phase} \\ \cmidrule(lr){2-3} \cmidrule(lr){4-5}
       & MSH & MSG  & MSH& MSG\\ 
\hline
    learning rate & $10^{-4}$ & $10^{-4}$  & $5 \times 10^{-4}$ & $5 \times 10^{-4}$\\
    batch size & 2000 &  2000  & 200 & 200\\
    Influence weight& 1.0& 1.0 & 1.0 & 1.0\\
    MOA loss weight&  3.0&  3.0 &  10.0 & 10.0\\
    entropy coefficient & 0.01 & 0.01 & 0.01& 0.01\\
\hline
\end{tabular}
}

\subtable[PR2]{
\renewcommand\arraystretch { 1.0 }
\centering
\begin{tabular}{c c c c c}
\hline 
& \multicolumn{2}{c}{self-play phase} & \multicolumn{2}{c}{adaptation phase} \\ \cmidrule(lr){2-3} \cmidrule(lr){4-5}
       & MSH & MSG  & MSH& MSG\\ 
\hline
    learning rate & $10^{-4}$ & $10^{-4}$  & $5 \times 10^{-4}$ & $5 \times 10^{-4}$\\
    batch size & 2000 &  2000  & 200 & 200\\
    soft update parameter& 0.99& 0.99 & 0.99 & 0.99\\
\hline
\end{tabular}
}

\end{table}

\section{Supplementary Results}

\subsection{Orcale Agents}
\label{sec: long interaction}
To compare and evaluate the performance of few-shot adaptation between HOP and learning baselines, we train an Orcale agent to see how well a well-established RL agent can perform in adaptation to co-players through extensive interactions 

Specifically, for every type of co-players, one Orcale agent interacts with them and is trained via A3C to converge from scratch.
During the training phase, co-players' parameters are fixed, which are the convergent parameters in their self-play.
In the subsequent adaptation phase, the trained Orcale agent is tested in the same way as HOP and baseline algorithms.
This process ensures that the Orcale agent engages in extensive interactions with the agents it would encounter during the adaptation phase.
Over an extended duration of interaction, Orcale effectively acquires a robust and high-quality policy.
We use the Orcale agent's performance in the adaptation phase as a reference point to explain HOP's performance.

\begin{table}[t]
\caption{Few-shot adaptation performance of Orcale in all three sequential social dilemma paradigms. The interaction happens between 1 Orcale agent and 3 co-players using the column policy. Shown is the average reward for Orcale from $1800$ to $2400$ step.}
\centering
\begin{tabular}{c c c c c c c}
    \hline
        & \multicolumn{3}{c}{learning co-players} & \multicolumn{3}{c}{rule-based co-players}\\ \cmidrule(lr){2-4} \cmidrule(lr){5-7}
        & LOLA &A3C & PS-A3C & random & cooperator & defector \\ 
    \hline
        MSH-4h1s& 2.44\err{0.03}  &  0.88\err{0.01} & 3.57\err{0.03} &1.10\err{0.00} & 2.73\err{0.02} & 0.93\err{0.01}\\ 
        MSH-4h2s & 3.23\err{0.02} & 3.46\err{0.01}& 3.97\err{0.02} &1.22\err{0.01} & 3.42\err{0.02} & 0.70\err{0.01}\\ 
        MSG & 20.9\err{0.12} & 22.7\err{0.17} & 32.5\err{0.12} & 16.0\err{0.08} & 36.0\err{0.00} & 12.0\err{0.00}\\ 
    \hline
    \end{tabular}
\label{table: LI-Ref}
\end{table}

\subsection{Ablation Study}
\label{sec: ablation study}

To test the importance and necessity of each component in HOP, we construct three partially ablated versions of HOP. The agent without inter-OM (w/o inter-OM) does not execute the inter-episode update expressed as \cref{eq: inter}. W/o inter-OM begins each episode with a uniform belief prior. The agent without intra-OM (w/o intra-OM) does not execute the intra-episode update expressed as \cref{eq: intra}. That is, for w/o intra-OM, $b_{ij}^{K, t}(g_j) = b_{ij}^{K, 0}(g_j), \forall t$. The direct-OM agent removes the whole opponent modeling module of HOP, and utilizes neural networks to model co-players directly. The co-player policies are mappings from states to actions, and not conditioned on goals. Experimental results for HOP and its three ablation versions in MSH-4h2s are shown in \cref{table: ablation}.

\begin{table}[ht]
\centering
\caption{Performance of HOP and its ablation versions in MSH-4h2s. In (a) self-play, 4 agents of the same kind are trained to converge. Shown is the normalized score after convergence. In (b) few-shot adaptation, the interaction happens between 1 agent using the row policy and 3 co-players using the column policy. Shown are the min-max normalized scores, with normalization bounds set by the rewards of Orcale and the random policy. The results are depicted for the row policy from 1800 to 2400 step.}

\subtable[Self-play performance]{
\centering
\begin{tabular}{c c c c}
\hline
 HOP & w/o inter-OM & w/o intra-OM & direct-OM \\ \hline
 \textbf{0.9767}\err{0.0117} & 0.9708\err{0.0146} & 0.9738\err{0.0117} & 0.9417\err{0.0146}\\
\hline
\end{tabular}
\label{subtable: ablation-self-play}
}

\subtable[Few-shot adaptation performance]{
\centering
\begin{tabular}{c c c c c c c}
\hline
        & \multicolumn{3}{c}{learning co-players} & \multicolumn{3}{c}{rule-based co-players}\\ \cmidrule(lr){2-4} \cmidrule(lr){5-7}
        & LOLA & A3C & PS-A3C & random & cooperator & defector\\ 
\hline
HOP & \textbf{0.97}\err{0.02} &\textbf{0.99}\err{0.02} & \textbf{0.88}\err{0.02} & \textbf{0.78}\err{0.07} & \textbf{1.00}\err{0.01} & \textbf{0.36}\err{0.02} \\
w/o inter-OM & \textbf{0.97}\err{0.02} & 0.92\err{0.03} & 0.87\err{0.02} & \textbf{0.78}\err{0.03} & 0.96\err{0.02} & 0.31\err{0.02} \\
w/o intra-OM & 0.95\err{0.02} & 0.98\err{0.02} & 0.84\err{0.01} & 0.65\err{0.04} & 0.99\err{0.02} & 0.34\err{0.03} \\
direct-OM & 0.95\err{0.01} &0.85\err{0.02} & 0.74\err{0.03} & 0.62\err{0.04} & 0.96\err{0.02} & 0.31\err{0.02} \\
\hline
\end{tabular}
\label{subtable:ablation-adaptation}
}
\label{table: ablation}
\end{table}

In self-play, HOP have an advantage over direct-OM agents. It suggests that utilizing a goal as a high-level representation of agents' behavior is beneficial to opponent modeling in complex environments. On the other hand, compared with w/o inter-OM and w/o intra-OM, HOP does not exhibit a significant advantage in self-play. The inter-OM and intra-OM modules may not be effective in the self-play setting, where a large number of interactions happen.

In the experiments testing few-shot adaptation, HOP outperforms its ablation versions. W/o inter-OM agents struggle when facing agents with fixed goals, such as cooperators and defectors. As the goals of cooperators and defectors are fixed, correct actions can be taken immediately if the focal agent has accurate goal priors. W/o inter-OM agents lack accurate goal priors at the beginning of an episode. In every episode, they have to use multiple interactions to infer co-players' goals and thus miss out on early opportunities to maximize their interests. 

W/o intra-OM agents exhibit poor performance when facing agents with dynamic behavior such as LOLA, PS-A3C, and random. These co-players have multiple goals. But in a given episode, the specific goals of a co-player can be gradually determined by analyzing its trajectory in this episode. However, w/o intra-OM agents can only count on inter-OM, which only takes the past episodes into account, but does not consider the information from the current episode. It results in inaccurate goal estimates in a given episode, which hurts the performance in few-shot adaptation.

Direct-OM agents are at an overall disadvantage. Their opponent modeling solely relies on the neural network, which makes it challenging to obtain significant updates during a short interaction. This leads to inaccurate opponent modeling during the adaptation phase. Furthermore, direct-OM agents utilize end-to-end opponent modeling, which introduces a higher degree of uncertainty compared to the goal-conditioned policy. This uncertainty can reduce the precision of the simulated co-player behavior during planning.

\section{Emergence of Social Intelligences}
\label{sec:emergence}
There are two kinds of social intelligence, self-organized cooperation and the alliance of the disadvantaged, emerging from the interaction between multiple HOP agents in MSH.
We make a minor modification to the game: the game terminates only when the time $T_{max}=30$ runs out.

\begin{figure}[htbp]
\centering
\subfigure[Self-organized cooperation]{
\includegraphics[width=0.8\linewidth]{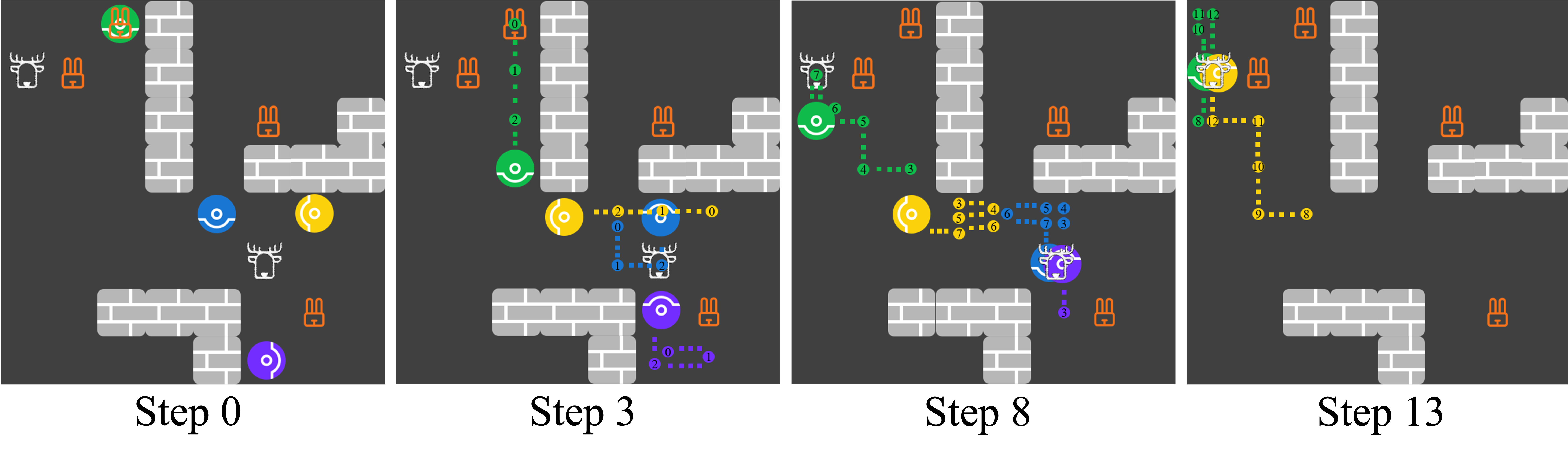}
\label{self-organization}
}%

\subfigure[Alliance of the disadvantaged]{
\includegraphics[width=0.8\linewidth]{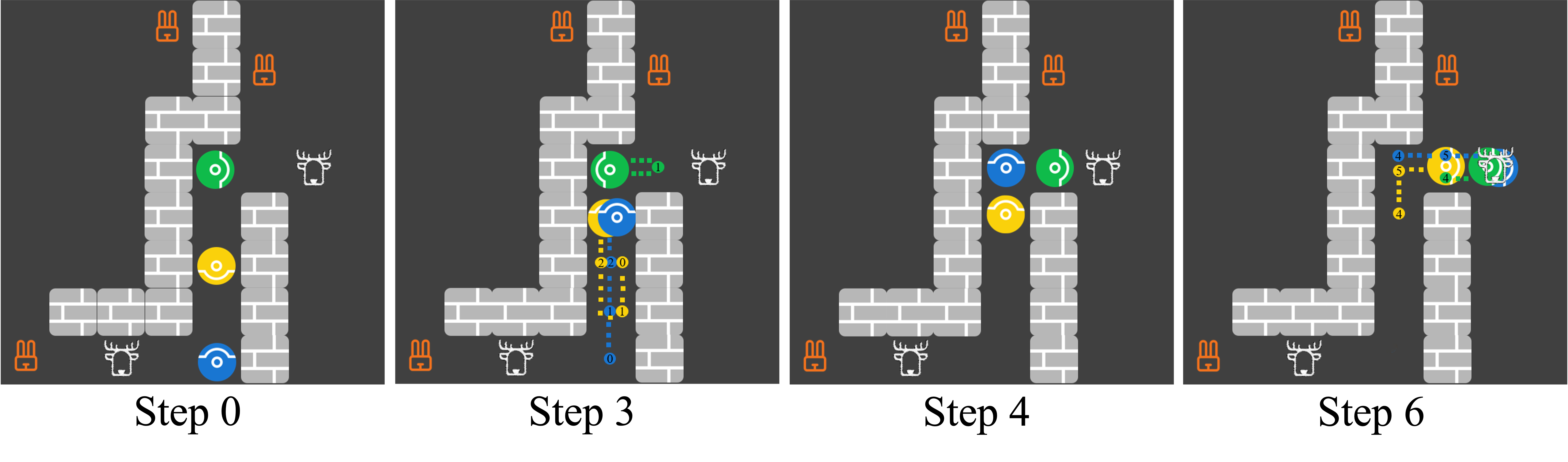}
\label{deception}
}%
\centering
\caption{Screenshots for the emergence of (a) self-organized cooperation and (b) alliance of the disadvantaged. Each panel shows agents' locations at the current step and the trajectories between the current step and the previously stated step.}
\end{figure}

\paragraph{Self-organized cooperation.} As shown in \cref{self-organization}, at the start of the game, three agents (blue, yellow, and purple) are two steps away from the stag at the bottom-right side, and the last agent (green) is spawned alone in the upper left corner.
One simple strategy for the three agents located at the bottom-right corner is to hunt the nearby stag together. 
Although this is a riskless strategy, the three agents each only obtain a reward of $10/3$. Instead, if one agent chooses to collaborate with the green agent at the top-left corner, all four agents each get a reward of $5$. 
This strategy is riskier since if the green agent chooses to hunt a nearby hare, the collaborative agent will not be able to catch any stag.
We show that HOP is able to achieve the aforementioned risky but rewarding collective strategy.
Specifically, the green agent refuses to catch the hare at his feet and shows the intention of cooperating with others (see screenshots at step 3 and step 8 in \cref{self-organization}). The yellow agent refuses to catch the stag at the bottom-right corner and chooses to collaborate with the green agent to hunt the stag in the top-left corner. In this process, all four agents receive the maximum profit.
Here, agents achieve pairwise cooperation through independent decision-making, without centralized assignment of goals.
Thus, we call this phenomenon self-organized cooperation.

\paragraph{Alliance of the disadvantaged.} In addition to the aforementioned game rules, we assume agents are heterogeneous.
Specifically, the yellow agent (Y) is three times greedier than the blue agent (B) and the green agent (G).
That is, when the three agents cooperate to hunt a stag successfully, Y will get a reward of $6$, and the others get $2$ each.
When Y cooperates with one of B and G, Y will obtain $7.5$, the other one gets $2.5$. 
As shown in \cref{deception}, at the start of the game, Y locates between B and G. 
Neither B nor G would like to cooperate with Y.
Hence they need to move past Y to cooperate with each other.
To achieve this, agents B and G first move closer to each other in the first few steps. However, to maximize its own profit, agent Y also moves toward B and G and hopes to hunt a stag with them.
To avoid collaboration with agent Y, after agents B and G are close enough to each other, they move back and forth to mislead Y (see step 3 of \cref{deception}).
Once agent Y makes a wrong guess of the directions agents B and G move, B and G will get rid of Y, and move to the nearest stag to achieve cooperation (see Step 4 and 6 of \cref{deception}), which maximizes the profit of agents B and G.

From the above two cases, we find that although HOP aims to maximize self-interest, cooperation emerges from the interaction between multiple HOP agents in mixed-motive environments.
This shows that it may be helpful in solving mixed-motive environments by equipping agents with the ability to infer others' goals and behavior and the ability to fast adjust their own responses.

\end{document}